\documentclass[11pt]{article}

\usepackage{graphicx}
\usepackage{amssymb}
\usepackage{subcaption}
\graphicspath{{pics/}}
\usepackage{soul}
\usepackage{xcolor}
\usepackage{bm}
\usepackage{amsmath}
\usepackage{amsthm}

\usepackage[linesnumbered,ruled,vlined]{algorithm2e}
\usepackage[export]{adjustbox}
\usepackage{natbib}
\usepackage{authblk}
\usepackage{hyperref}

\usepackage[utf8]{inputenc}
\newtheorem{definition}{Definition}[section]
\newtheorem{theorem}{Theorem}[section]
\usepackage[section]{placeins}
\theoremstyle{definition}
\newtheorem{definition1}{Definition}[section]
\usepackage{tabularx}

\usepackage{geometry}
\title{Gradient-only line searches: An Alternative to Probabilistic Line Searches}

\author[1]{Dominic Kafka}
\author[1]{Daniel N. Wilke}
\affil[1]{Centre for Asset and Integrity Management (C-AIM),
	Department of Mechanical and Aeronautical Engineering,
	University of Pretoria, Pretoria, South Africa}
\affil[ ]{\textit{\{dominic.kafka@gmail.com, wilkedn@gmail.com\}}}

\begin{document}
	
\maketitle

\begin{abstract}%
Step sizes in neural network training are largely determined using predetermined rules such as fixed learning rates and learning rate schedules. These require user input or expensive global optimization strategies to determine their functional form and associated hyperparameters. Line searches are capable of adaptively resolving learning rate schedules. However, due to discontinuities induced by mini-batch sub-sampling, they have largely fallen out of favour. Notwithstanding, probabilistic line searches, which use statistical surrogates over a limited spatial domain, have recently demonstrated viability in resolving learning rates for stochastic loss functions. 

This paper introduces an alternative paradigm, Gradient-Only Line Searches that are Inexact (GOLS-I), as an alternative strategy to automatically determine learning rates in stochastic loss functions over a range of 15 orders of magnitude without the use of surrogates. We show that GOLS-I is a competitive strategy to reliably determine step sizes, adding high value in terms of performance, while being easy to implement.



\end{abstract}

{\bf Keywords:} Artificial Neural Networks, Gradient-only, Line Searches, Learning Rates, Discontinuities, SNN-GPP

\section{Introduction}
\label{Introduction}

\begin{figure}[ht]
	\centering
	\includegraphics[scale=0.4]{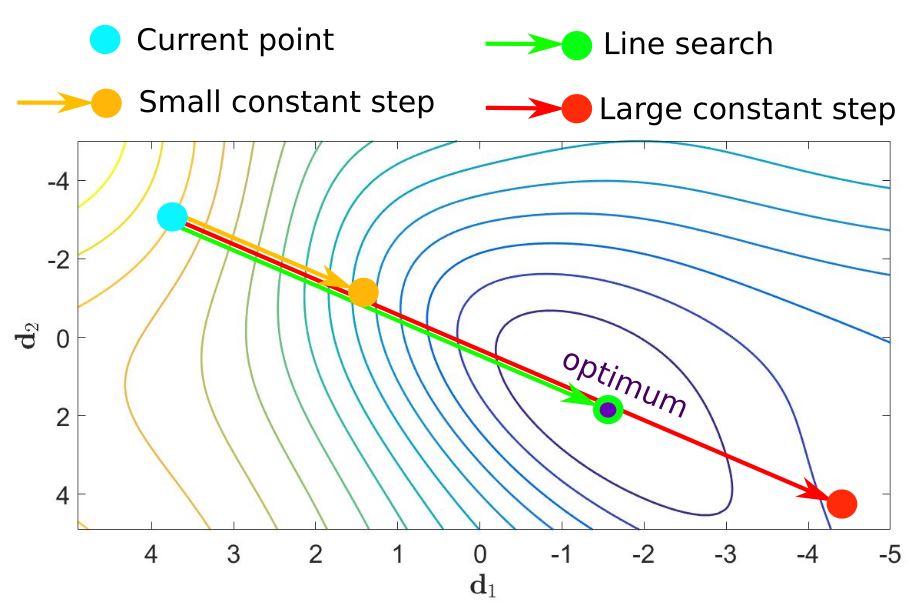}
	\caption{Contour plot of a neural network loss along two random perpendicular directions. Finding an optimum along a search direction: Too small constant steps are expensive, while too large constant steps can be unstable during training. Line searches balance performance and stability.}
	\label{fig_intro_LS}
\end{figure}

\begin{figure}[ht]
	\centering
	\includegraphics[scale=0.45]{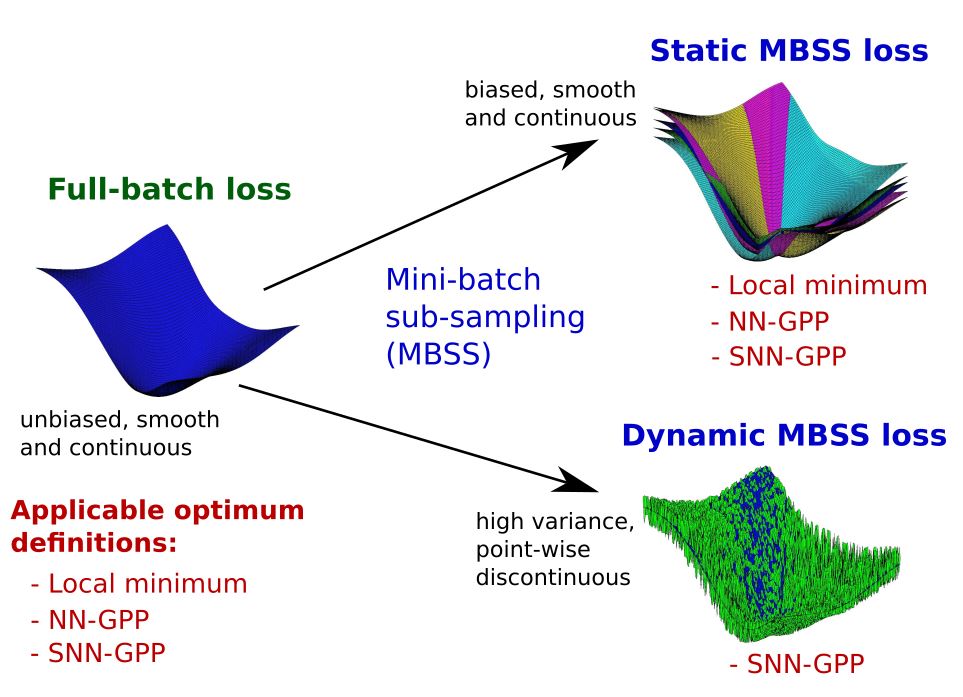}
	\caption{Full and reduced fidelity loss representations with various definitions used to identify optima. Mini-batch sub-sampling (MBSS) can be conducted using static and dynamic approaches, resulting in smooth and point-wise discontinuous loss approximations respectively. The local minimum, non-negative associated gradient projection point (NN-GPP) and stochastic NN-GPP (SNN-GPP) \citep{Kafka2019jogo} definitions apply to full-batch and static MBSS losses. However, only the SNN-GPP is effective in localizing optima in dynamic MBSS loss functions.}
	\label{fig_intro_summary}
\end{figure}

Determining the learning rate, or learning rate schedule parameters, is still an active field of research in deep learning \citep{Smith2015,Orabona2017,Wu2018}, as these parameters have been shown to be the most sensitive hyperparameters in neural network training \citep{Bergstra2012}. In practice, their magnitudes are often selected {\it a priori} by the user. If these parameters result in update steps that are too small, training is stable, but computationally expensive. Conversely, with updates that are too large, training becomes unstable. In mathematical programming learning rates (step sizes) are commonly automatically resolved by line searches \citep{Arora2011}, see Figure~\ref{fig_intro_LS}. However, these traditionally require smooth and continuous loss functions on which to operate, see the full-batch loss in Figure~\ref{fig_intro_summary}.

In modern deep learning tasks the dataset sizes exceed the memory capabilities of individual computational nodes. In particular with the rise of memory-limited parallel computing platforms such as graphical processing units (GPUs) in deep learning, it is infeasible to evaluate full-batch loss functions. Instead, a mini-batch of available training data is sub-sampled to evaluate an approximate loss function. In fields such as adaptive sub-sampling methods, the primary concern is to resolve approximate loss functions with desired characteristics. The aim might be to select a sub-sample that results in the best mini-batch approximation of the full-batch loss function, or to ensure that selected approximate losses on average result in descent directions, \citep{Friedlander2011,Bollapragada2017}. This means that for a standard update step, which consists of a search direction and corresponding step size, the focus of adaptive sampling methods is primarily on solving for the quality of the search direction. To make the most of the carefully selected mini-batch, it is kept constant while conducting a line search along the given search direction \citep{Martens2010,Friedlander2011,Byrd2011,Byrd2012,Bollapragada2017,Kungurtsev2018,Paquette2018,Bergou2018,Mutschler2019}. We call this approach static mini-batch sub-sampling (MBSS), see Figure~\ref{fig_intro_summary}, which constructs different loss function surfaces for every static mini-batch. Critically, this means that the loss function presented to a line search is continuous and smooth, which allows for the use of minimization line searches. However, the act of fixing the mini-batch introduces a sampling error, which biases the step size to the given mini-batch. The consequence is, that a different mini-batch can lead a line search to finding an alternative minimum, as demonstrated by the cyan surface in Figure~\ref{fig_intro_summary}.

An alternative is to continuously resample new data for every approximate loss evaluation\citep{Mahsereci2017a,Wills2018,Wills2019}. This approach was first introduced within the context of line searches by \cite{Mahsereci2017a}, though at the time not explicitly differentiated from static MBSS as used in adaptive sampling methods. We call the method of repeatedly sampling a new mini-batch for every function evaluation along a search direction {\it dynamic MBSS}, also known as approximate optimization \citep{Bottou2010}. However, this spoils the utility of minimization line searches in neural network training, as randomly alternating between the sampling error associated with each mini-batch introduces discontinuities in resulting loss functions and gradients, see Figure~\ref{fig_intro_summary}. The consequence is that alternating between approximate loss expressions, critical points may not exist, and line searches falsely identify discontinuities as local minima \citep{Wilson2003,Schraudolph2003,Schraudolph2006}. This led to line searches being replaced by {\it a priori} rule based step size schedules typical of sub-gradient methods, including stochastic gradient descent (SGD) \citep{Schraudolph1999,Boyd2014,Smith2015}.

Recently, the use of Gaussian process surrogates using both function value and gradient information along search directions has reintroduced line searches to neural network training \citep{Mahsereci2017a}. However, we postulate that a simpler and more accessible approach may be sufficient to construct line searches, using only gradient information. The premise for this proposition is that 1) a stochastic adaptation to the gradient-only optimality formulation of the Non-Negative Associated Gradient Projection Point (NN-GPP) \citep{Wilke2013,Snyman2018}, i.e. the stochastic NN-GPP (SNN-GPP) \citep{Kafka2019jogo}, is superior to local minima in identifying true optima in stochastic loss functions and 2) the discontinuities in function values are more severe than those in directional derivatives \citep{Mahsereci2017a,Kafka2019b}, aiding SNN-GPP identification. 

In this paper we demonstrate how the combination of these characteristics allows for the construction of gradient-only line searches that automatically determine step sizes. Based on empirical evidence, we argue that developing line searches that locate SNN-GPPs offers a competitive, light-weight and more robust alternative to conducting probabilistic line searches. We also present three examples of using gradient-only line searches as a research tool.

\subsection{Our Contribution}

In this paper, we compare the {\it Gradient-Only Line Search that is Inexact} (GOLS-I) \citep{Kafka2019jogo} to the probabilistic line search (PrLS) proposed by \cite{Mahsereci2017a} for determining learning rates in dynamically mini-batch sub-sampled loss functions. GOLS-I approximates the location of Stochastic Non-Negative Associated Gradient Projection Points (SNN-GPP) \citep{Kafka2019jogo}, an adaptation of the gradient-only optimality criterion \citep{Wilke2013,Snyman2018}. When considering univariate functions, such as loss functions along a search direction, a SNN-GPP manifests as a sign change from negative to positive in the directional derivative along the descent direction. We stress that we do not rely on the concept of a critical point \citep{Arora2011} as we do not require the derivative at a SNN-GPP to be zero. Specifying a sign change from negative to positive and not from positive to negative along a descent direction, incorporates second order information in that it reflects a local minimum as determined by gradient information.

Some common learning rate schedules use step sizes ranging over 5 orders of magnitude \citep{Senior2013}, while the magnitudes of cyclical learning rate schedules typically range over 3 to 4 orders of magnitude \citep{Smith2015,Loshchilov2016}. Manually selected schedules can require a number of hyperparameters to be determined. Our proposed method, GOLS-I, can determine step sizes over a range of 15 orders of magnitude without the need for any parameter tuning. The high range of available step sizes within the line search allow GOLS-I to effectively traverse flat planes or steep declines in discontinuous stochastic loss functions, while requiring no user intervention.

We also use this platform to 1) explicitly compare static and dynamic MBSS in the context of line searches, 2) uncouple the quality of search directions from the accuracy of resolving optima along the given direction and 3) demonstrate the sensitivity of search directions in SGD to mini-batch size in comparison to full-batch descent directions.



\section{Loss Functions and means of locating optima}

Commonly, the loss functions used in neural network training have the form
\begin{eqnarray}
\mathcal{L}(\boldsymbol{x}) = \frac{1}{M} \sum_{b=1}^{M} \ell (\boldsymbol{x};\;\boldsymbol{t}_b),
\label{eq:loss}
\end{eqnarray}

where $\{\boldsymbol{t}_1,\dots,\boldsymbol{t}_M\}$ is a training dataset of size $M$, $\boldsymbol{x}\in \mathcal{R}^p$ is an $p$-dimensional vector of model parameters, and $\ell(\boldsymbol{x};\;\boldsymbol{t})$ defines the loss quantifying the fitness of
parameters $\boldsymbol{x}$ with regards to training sample $\boldsymbol{t}$. Backpropagation \citep{Werbos1994} computes the exact gradient w.r.t.  $\boldsymbol{x}$, resulting in:
\begin{eqnarray}
\nabla\mathcal{L}(\boldsymbol{x}) = \frac{1}{M} \sum_{b=1}^{M} \nabla\ell (\boldsymbol{x};\;\boldsymbol{t}_b).
\label{eq:lossgrad}
\end{eqnarray}

In this case, all the training data is used for both function and gradient evaluations, resulting in the true or full-batch loss, $ \mathcal{L}(\boldsymbol{x}) $, and gradients, $ \nabla\mathcal{L}(\boldsymbol{x}) $, which are continuous and smooth. However, in modern deep learning problems the cost of computing the full-batch loss, $\mathcal{L}(\boldsymbol{x})$, is high. Therefore, mini-batch sub-sampling (MBSS) can be introduced to generate loss function approximations with using a subset of the training data, $\mathcal{B} \subset \{1,\dots,M\}$ of size $|\mathcal{B}| \ll M$, resulting in:

\begin{equation}
L(\boldsymbol{x}) = \frac{1}{|\mathcal{B}|} \sum_{b\in \mathcal{B}} \ell (\boldsymbol{x};\;\boldsymbol{t}_b),
\label{eq:lossgradbatch}
\end{equation}
and corresponding approximate gradient
\begin{equation}
\boldsymbol{g}(\boldsymbol{x}) = \frac{1}{|\mathcal{B}|} \sum_{b\in \mathcal{B}} \nabla\ell (\boldsymbol{x};\;\boldsymbol{t}_b).
\label{eq:g_lossgradbatch}
\end{equation}

The approximate loss function has expectation $ E [ L(\boldsymbol{x}) ] = \mathcal{L}(\boldsymbol{x}) $ and corresponding expected gradient $ E [ 
\boldsymbol{g}(\boldsymbol{x}) ] = \nabla\mathcal{L}(\boldsymbol{x})$ \citep{Tong2005}, but individual instances may vary significantly from the mean. Evaluating $L(\boldsymbol{x})$ instead of $\mathcal{L}(\boldsymbol{x})$ decreases the computational cost and increases the chance of an optimization algorithm overcoming local minima. 

To formally extend our discussion to line searches, consider the following notation: We define a univariate function at given iteration $n$ of stochastic gradient descent that uses a line search (LS-SGD) \cite{Kafka2019jogo} as $F_n(\alpha)$ along a descent direction, $\boldsymbol{d}_n\in\mathcal{R}^p$ from $\boldsymbol{x}_n\in\mathcal{R}^p$: 
\begin{equation}
F_n(\alpha) = f(\boldsymbol{x}_n(\alpha)) = L(\boldsymbol{x}_n + \alpha \boldsymbol{d}_n),
\label{eq_linesearch}
\end{equation}
with associated directional derivative
\begin{equation}
F_n'(\alpha) = \frac{d F_n(\alpha)}{d \alpha} = \boldsymbol{d}_n \cdot \boldsymbol{g}(\boldsymbol{x}_{n} + \alpha \boldsymbol{d}_n ).
\label{eq_lineg}
\end{equation}

If full-batch sampling is implemented, the univariate loss representations along a search direction $\boldsymbol{d}_n$ is denoted $\mathcal{F}_n(\alpha)$ with respective derivative $\mathcal{F}_n(\alpha)$.
Suppose static mini-batch sub-sampling (MBSS) is implemented, where:
\begin{definition}
	\label{def_staticMBSS}
	Static mini-batch sub-sampling is conducted when the mini-batch, $\mathcal{B}$, used to evaluate approximations $L(\boldsymbol{x})$ and $\boldsymbol{g}(\boldsymbol{x})$ remains constant for a minimum duration of conducting a line search along a fixed search direction in iteration $n$ of a training algorithm. Therefore, mini-batches selected using static MBSS are denoted as $\mathcal{B}_n$. The overhead bar notation is used to identify approximations evaluated using static mini-batch sub-sampling as $\bar{L}(\boldsymbol{x})$ and $\bar{\boldsymbol{g}}(\boldsymbol{x})$ respectively.   \citep{Kafka2019jogo}	
\end{definition}
The static MBSS loss is substituted into Equations (\ref{eq_linesearch}) and (\ref{eq_lineg}) give static MBSS univariate functions $\bar{F}_n(\alpha)$ and $\bar{F}_n'(\alpha)$ respectively. 

\begin{figure}[h!]
	\centering
	\begin{subfigure}{.39\textwidth}
		\centering 
		\includegraphics[width=1\linewidth]{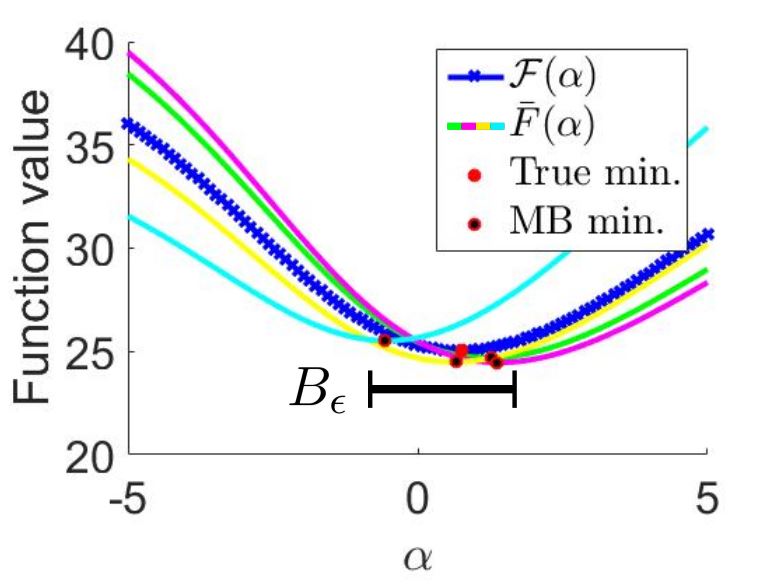}
		\caption{}
	\end{subfigure}%
	\begin{subfigure}{.39\textwidth}
		\centering
		\includegraphics[width=1\linewidth]{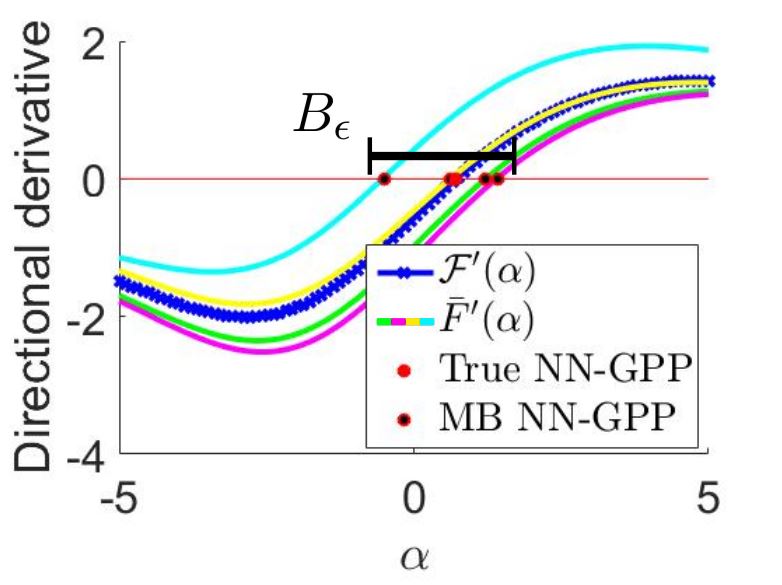}
		\caption{}
	\end{subfigure}%
	
	\begin{subfigure}{.39\textwidth}
		\centering 
		\includegraphics[width=1\linewidth]{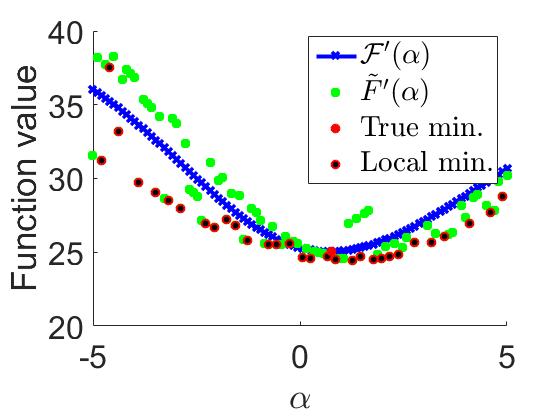}
		\caption{}
	\end{subfigure}%
	\begin{subfigure}{.39\textwidth}
		\centering
		\includegraphics[width=1\linewidth]{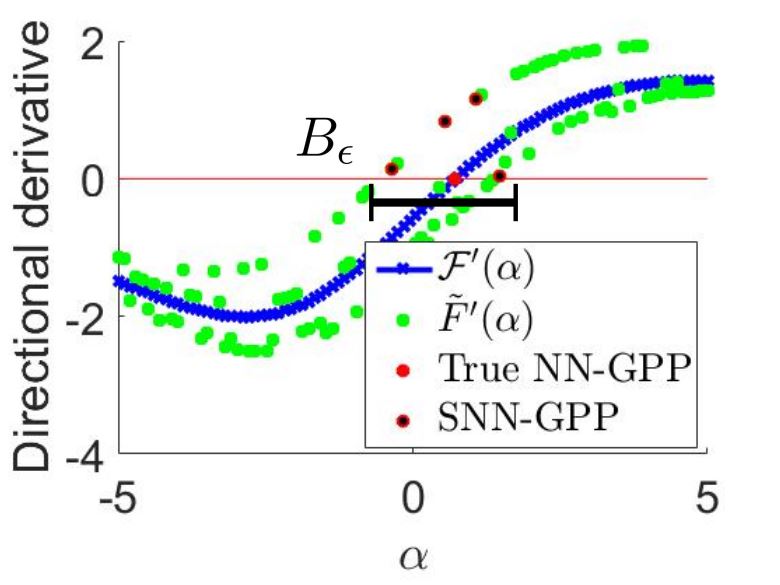}
		\caption{}
	\end{subfigure}%
	
	\begin{subfigure}{.39\textwidth}
		\centering 
		\includegraphics[width=1\linewidth]{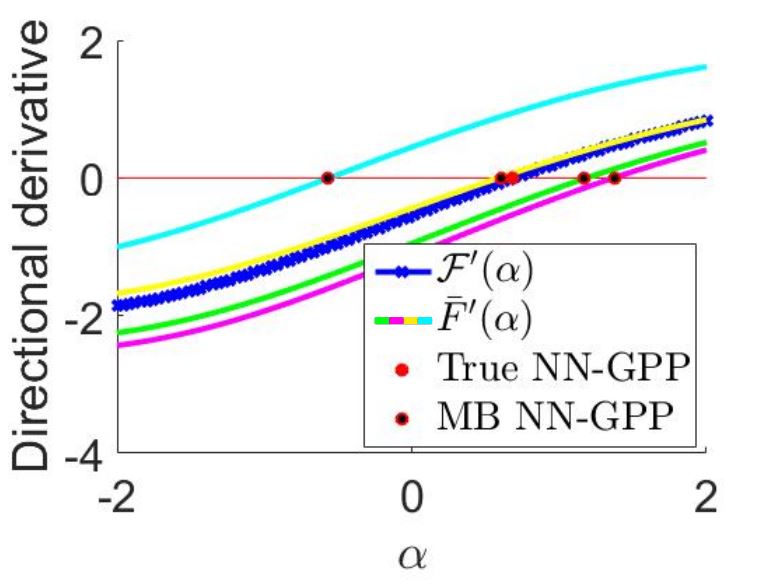}
		\caption{}
	\end{subfigure}%
	\begin{subfigure}{.39\textwidth}
		\centering
		\includegraphics[width=1\linewidth]{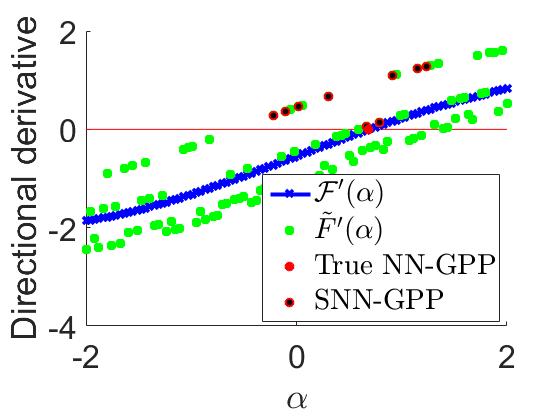}
		\caption{}
	\end{subfigure}%
	
	\caption{Comparing univariate loss functions and directional derivatives along a search direction with (a)(b) full-batch and static MBSS evaluated loss functions, and (c)(d) full-batch and dynamic MBSS loss functions. (e) and (f) show a closer comparison of the directional derivatives of both MBSS modes.	}
	\label{fig_stVdyn_1D}
\end{figure}

In Figure~\ref{fig_stVdyn_1D} we use a simple single hidden layer neural network with Sigmoid activation functions applied to the famous Iris \citep{Fisher1936} dataset as an example of a loss function used in neural network training. Note, that subscript $n$ is dropped, as the plots represent a single search direction. This allows us to explore loss function characteristics and applicable optimality definitions typically encountered during neural network training. Shown in blue in Figures~\ref{fig_stVdyn_1D}(a) and (b) are $\mathcal{F}(\alpha)$ and directional derivative $\mathcal{F}'(\alpha)$ for the full-batch loss evaluation of our illustrative problem along arbitrary search direction $\boldsymbol{d}$. Subsequently, the Iris training dataset is broken into 4 equal sized mini-batches, each resulting in a unique function of $\bar{F}(\alpha)$ and $\bar{F}'(\alpha)$, plotted in green, magenta, yellow and cyan respectively. 

Both $\mathcal{F}(\alpha)$ and $\bar{F}(\alpha)$ with respective derivatives are smooth and continuous, meaning that local minima and critical points are defined. Therefore, minimization line searches are effective in finding the optima of these functions. However, note how conducting static MBSS incorporates bias into the loss approximations. Due to the nature of using a sub-set of the full training data, a {\it sampling error} is present for each univariate loss, resulting in the presence of a different minimizer for each mini-batch. Consequently, we define a ball $B_\epsilon$ for the MBSS case, which contains the location of all possible optima due to different mini-batches. In the 1D case, this ball simplifies to a range.



Now, suppose that dynamic MBSS is implemented in our example problem, where:
\begin{definition}
	\label{def_dynamicMBSS}
	Dynamic mini-batch sub-sampling is conducted when the mini-batch, $\mathcal{B}$, used to evaluate approximations $L(\boldsymbol{x})$ and $\boldsymbol{g}(\boldsymbol{x})$ changes for every evaluation, $i$, of $L(\boldsymbol{x})$ and $\boldsymbol{g}(\boldsymbol{x})$ in a line search along a fixed search direction for iteration, $n$, of a training algorithm. Therefore, mini-batches selected using dynamic MBSS are denoted as $\mathcal{B}_{n,i}$. The overhead tilde notation is used to identify approximations evaluated using dynamic MBSS as $\tilde{L}(\boldsymbol{x})$ and $\tilde{\boldsymbol{g}}(\boldsymbol{x})$ respectively. The mini-batch used to evaluate a given instance of both $\tilde{L}(\boldsymbol{x})$ and $\tilde{\boldsymbol{g}}(\boldsymbol{x})$ is identical, but subsequent evaluations of the $\tilde{L}(\boldsymbol{x})$ and $\tilde{\boldsymbol{g}}(\boldsymbol{x})$ pair prompts the re-sampling of a new mini-batch, $\mathcal{B}_{n,i}$.
\end{definition}
The unimodal functions constructed by applying dynamic MBSS to $F(\alpha)$ and $F'(\alpha)$ are denoted $\tilde{F}(\alpha)$ and $\tilde{F}'(\alpha)$ respectively. In our Iris example in Figure~\ref{fig_stVdyn_1D}(c) and (d) we randomly select one of the four mini-batches (that result in cyan, yellow, green and magenta $\bar{F}(\alpha)$ curves) for every loss function evaluation, $i$. The resulting loss approximation is discontinuous in both function values and directional derivatives, as the sampling error constantly changes, depending on the mini-batch selected at the given function evaluation. This leads to local minima being identified at discontinuities over the whole sampled domain, see Figure~\ref{fig_stVdyn_1D}(c). These discontinuities also imply that the first order optimality criterion $\tilde{F}'(\alpha^*) = 0$ \citep{Arora2011} for a local minimum may not exist for a given instance of $\tilde{F}'(\alpha^*)$, see Figure~\ref{fig_stVdyn_1D}(d), even if it may exist for the full-batch case, $\mathcal{F}(\alpha^*) = 0$.

An alternative gradient-only optimality criterion for discontinuous functions exists, namely the non-negative associated gradient projection point (NN-GPP) \citep{Wilke2013,Snyman2018} given by:
\theoremstyle{definition}
\begin{definition1}{NN-GPP:}
	A non-negative associated gradient projection point (NN-GPP) is defined as any point, $\boldsymbol{x}_{nngpp}$, for which there exists
	$r_u > 0$ such that 
	\begin{equation}
	\nabla f(\boldsymbol{x}_{nngpp} + \lambda \boldsymbol{u} )\boldsymbol{u} \geq 0,\;\;  \forall\; \boldsymbol{u} \in \left\{\boldsymbol{y}\in\mathbb{R}^p \;|\; \| \boldsymbol{y} \|_2 = 1\right\},\;\;\forall \;\lambda \in (0,r_u].
	\end{equation}	
	\label{def_nngpp}
\end{definition1}

\begin{figure}[!h]
	\centering 
	\includegraphics[width=0.8\linewidth]{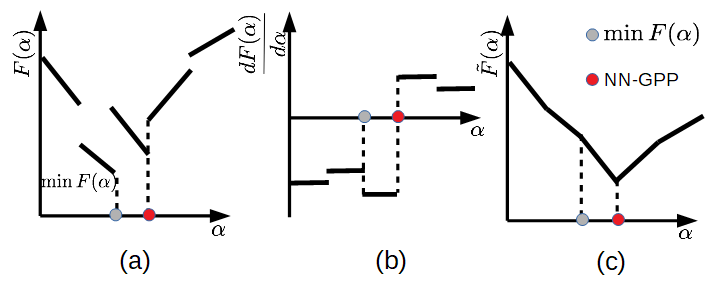}
	\caption{(a) Discontinuous stochastic function with (b) derivatives and (c) an alternative interpretation of (a) that is consistent with the associated derivatives given in (b). The function minimizer of $F(\alpha)$ (gray dot) and sign change from negative to positive along $\alpha$ (red dot) are indicated.}\label{fig:nngpp}
\end{figure}

This definition was proposed for deterministic static point-wise discontinuous functions, but generalizes to a minimum and/or a semi-definite critical point in smooth and continuous functions \citep{Wilke2013,Snyman2018}. The NN-GPP definition incorporates second order information in the form of requiring that within a certain radius, there are no descent directions away from a NN-GPP. The associated gradient \citep{Snyman2018} defines the  derivative at a discontinuity. Essentially, when we only consider associated derivatives along a line search we may interpret the discontinuous function presented in Figure~\ref{fig:nngpp}(a) to be equivalent to the continuous function presented in Figure~\ref{fig:nngpp}(c), since both are consistent with the associated derivatives presented in Figure~\ref{fig:nngpp}(b). The NN-GPP therefore filters out or ignores all discontinuities present in a discontinuous function. It is also clear, that the function minimizer of the discontinuous function, depicted as a grey dot in Figure~\ref{fig:nngpp}(a), is associated with a negative directional derivative in its neighbourhood along the direction $\alpha = +1$. This implies that the global function minimizer present in the discontinuous function is not representative of a local minimum according to the associated derivatives, \citep{Wilke2013,Snyman2018}. This is because a global or local minimum would be characterized by a directional derivative going from negative to positive along a descent direction. Traditionally, for $C^1$ smooth functions the derivative would be zero at the local minimum, indicative of a critical point \citep{Snyman2018}. As NN-GPPs were developed specifically for discontinuous functions, \citep{Wilke2013,Snyman2018}, it does not rely on the concept of a critical point as there is usually no point where the derivative is zero when discontinuous stochastic functions are considered. A NN-GPP along a search direction manifests as a sign change from negative to positive as along the descent direction. As outlined by \cite{Wilke2012}, this way of characterizing solutions of discontinuous stochastic functions is also consistent with solutions that sub-gradient algorithms or SGD (with constant step size) would find, i.e. using SGD to optimize Figure~\ref{fig:nngpp}(a) would only result in converge around the NN-GPP (red dot), while the global function minimizer (grey dot) would be ignored.

%

%

Returning to our Iris example in Figure~\ref{fig_stVdyn_1D}(b), the NN-GPP definition correctly identifies critical points for both the true, full-batch loss function $\mathcal{F}'(\alpha)$ (True NN-GPP) and the static MBSS loss approximations $\bar{F}'(\alpha)$ (MB NN-GPP). However, jumping between batches causes sign changes to occur where there are no NN-GPPs for any individual instance of $\bar{F}'(\alpha)$, see Figure~\ref{fig_stVdyn_1D}(e) and (f). Consider $B_\epsilon$, i.e. the range between the NN-GPP of the cyan $\bar{F}'(\alpha)$, and that of the magenta $\bar{F}'(\alpha)$ curve. Alternating between instances of $\tilde{F}'(\alpha)$ in $B_\epsilon$ will result in sign changes in the sampled directional derivatives, depending on the sequence of mini-batches chosen. The locations of these sign changes are unlikely to be representative of NN-GPPs. The definition of NN-GPPs were therefore extended for this stochastic setting to stochastic NN-GPPs \citep{Kafka2019jogo}:

\theoremstyle{definition}
\begin{definition1}{SNN-GPP:}
	A non-negative associated gradient projection point (SNN-GPP) is defined as any point, $\boldsymbol{x}_{snngpp}$, for which there exists
	$r_u > 0$ such that 
	$$
	\nabla f(\boldsymbol{x}_{snngpp} + \lambda \boldsymbol{u} )\boldsymbol{u} \geq 0,\;\;  \forall\; \boldsymbol{u} \in \left\{\boldsymbol{y}\in\mathbb{R}^p \;|\; \| \boldsymbol{y} \|_2 = 1\right\},\;\;\forall \;\lambda \in (0,r_u],
	$$
	with non-zero probability.
	\label{def_snngpp}
\end{definition1}

This definition caters for dynamic MBSS, incorporating the discontinuity-filtering nature of NN-GPPs outside of $B_\epsilon$, while accommodating sign changes in directional derivatives contained within $B_\epsilon$ that is not representative of a specific NN-GPP for any individual mini-batch. SNN-GPPs generalize to NN-GPPs, since a NN-GPP is a SNN-GPP with probability 1 for smooth and continuous functions. However, in the case of dynamic MBSS, even if $\alpha = \alpha_{nngpp}$, the probability of encountering a SNN-GPP is $<1$ over a large number of sub-samples at $\alpha$. This probability depends on the variance in $\tilde{\boldsymbol{g}}$, which is a function of $|\mathcal{B}_{n,i}|$ and the sampling strategy used. In our practical implementations, we limit ourselves to sampling mini-batches, $\mathcal{B}_{n,i}$, uniformly with replacement, keeping the mini-batch size, $|\mathcal{B}_{n,i}|$ constant. Subsequently, the number of possible mini-batch combinations is $K = \binom{M}{|\mathcal{B}_{n,i}|}$. When $M$ is large, the number of combinations, $K$ is so large, that discontinuities are often interpreted as stochastic noise \cite{Simsekli2019a}. Though this interpretation is undoubtedly useful, we suggest that strictly speaking, dynamic MBSS loss functions are discontinuous, where the magnitude of discontinuities depends on $M$, $|\mathcal{B}_{n,i}|$, the sampling strategy used and the data itself.

\section{Empirical evidence of improved localization of optima with SNN-GPPs over minimizers}

\begin{figure}[ht]
	\centering
	\begin{subfigure}{.5\textwidth}
		\centering 
		\includegraphics[width=1\linewidth]{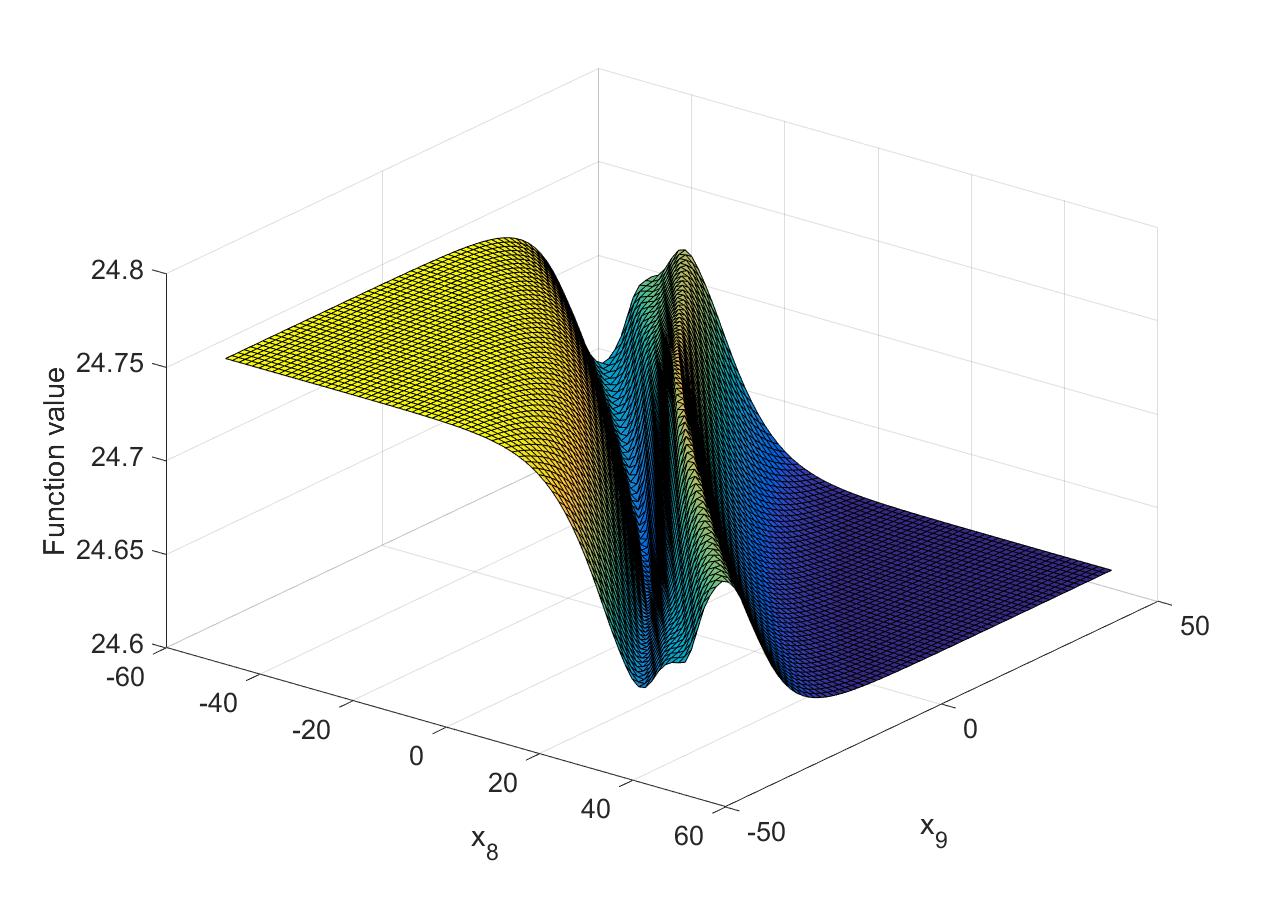}
		\caption{}
		\label{fig_cfunc_full_f}
	\end{subfigure}%
	\begin{subfigure}{.5\textwidth}
		\centering
		\includegraphics[width=1\linewidth]{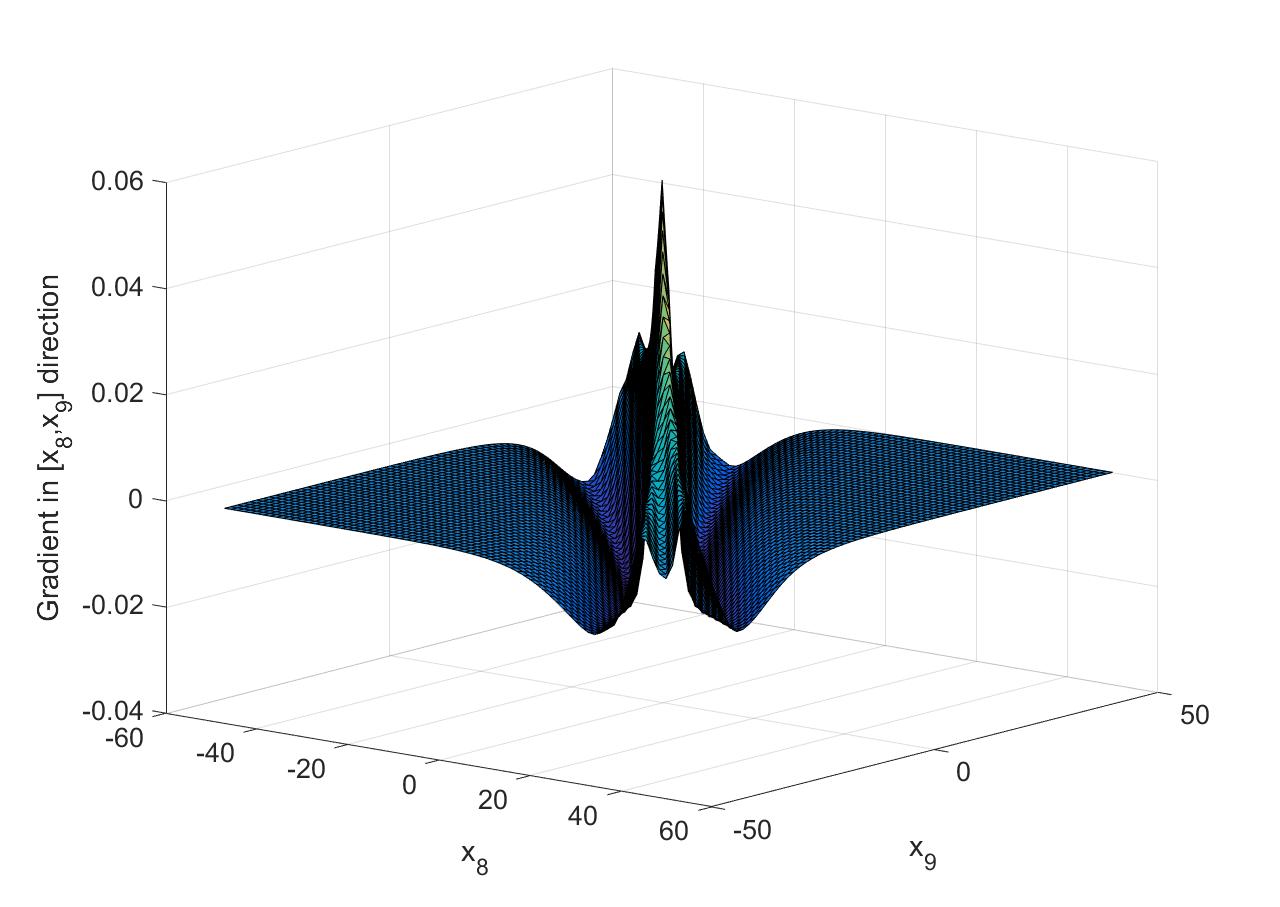}
		\caption{}
		\label{fig_cfunc_full_g}
	\end{subfigure}%
	
	\begin{subfigure}{.5\textwidth}
		\centering 
		\includegraphics[width=1\linewidth]{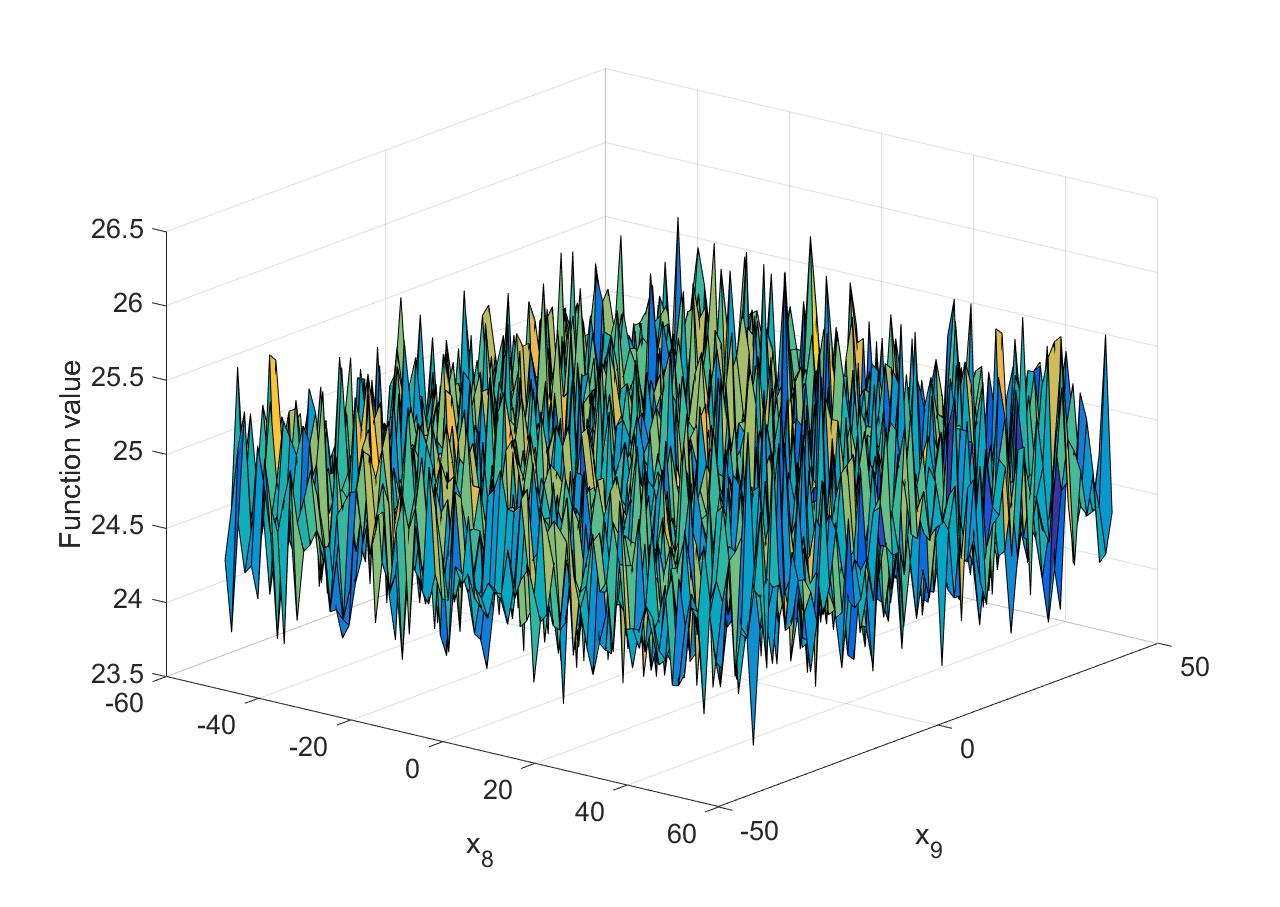}
		\caption{}
		\label{fig_cfunc_s10_f}
	\end{subfigure}%
	\begin{subfigure}{.5\textwidth}
		\centering
		\includegraphics[width=1\linewidth]{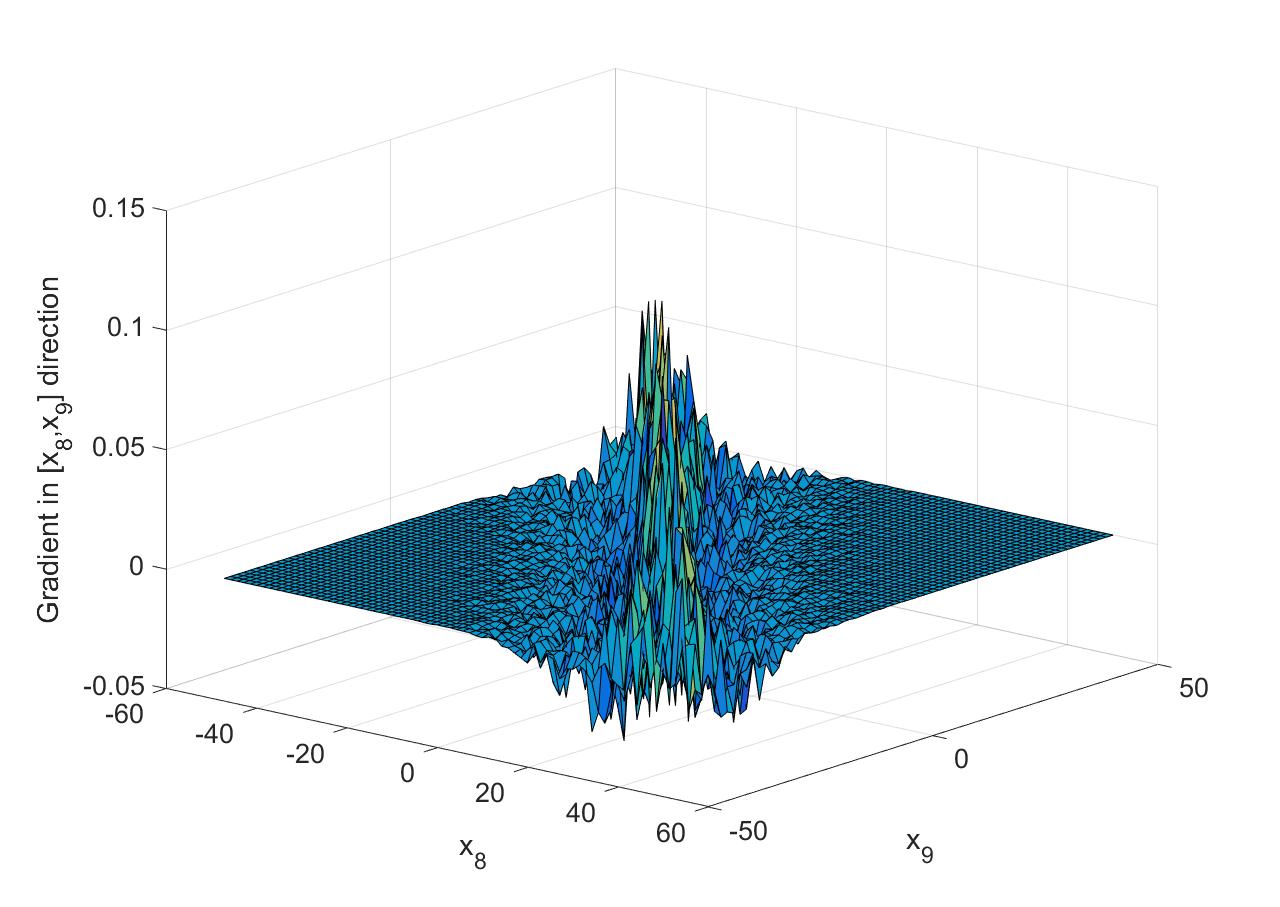}
		\caption{}
		\label{fig_cfunc_s10_g}
	\end{subfigure}%

	\caption{(a) Function values and (b) the directional derivatives of the loss function in dimensions $x_8$ and $x_9$ for a single hidden layer neural network applied to the Iris dataset \citep{Fisher1936}. Directional derivatives are generated using a fixed search direction $\boldsymbol{d}$, where the only non-zero components are $x_8$ and $x_9$, equal to $\frac{1}{\sqrt{2}}$. The directional derivative is then evaluated as $ \frac{d F_n}{d \alpha} = \boldsymbol{g}(\boldsymbol{x}_n) \cdot \boldsymbol{d} $, to generate the plots. When using full batches, both the function value and the directional derivatives are smooth and continuous functions.
		(c) Function values and (d) directional derivatives are discontinuous, when dynamic MBSS with mini-batch size $|\mathcal{B}_{n,i}| = 10$ is implemented. The function value plot's shape is not recognizable in comparison to (a), while directional derivatives still contain features of the original shape.}
	\label{fig_cfunc_full}
\end{figure}

\begin{figure}[h!]
	\centering
	\begin{subfigure}{.5\textwidth}
		\centering 
		\includegraphics[width=0.9\linewidth]{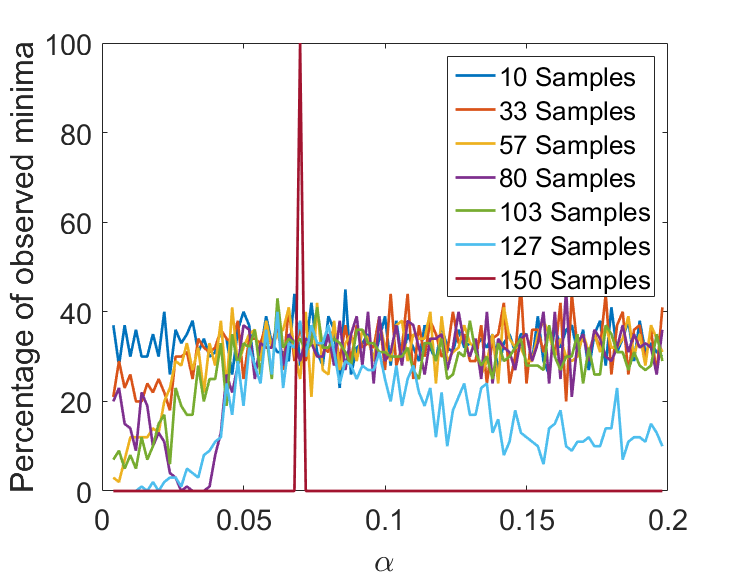}
		\caption{}
		\label{fig_dists_d11_f}
	\end{subfigure}%
	\begin{subfigure}{.5\textwidth}
		\centering
		\includegraphics[width=0.9\linewidth]{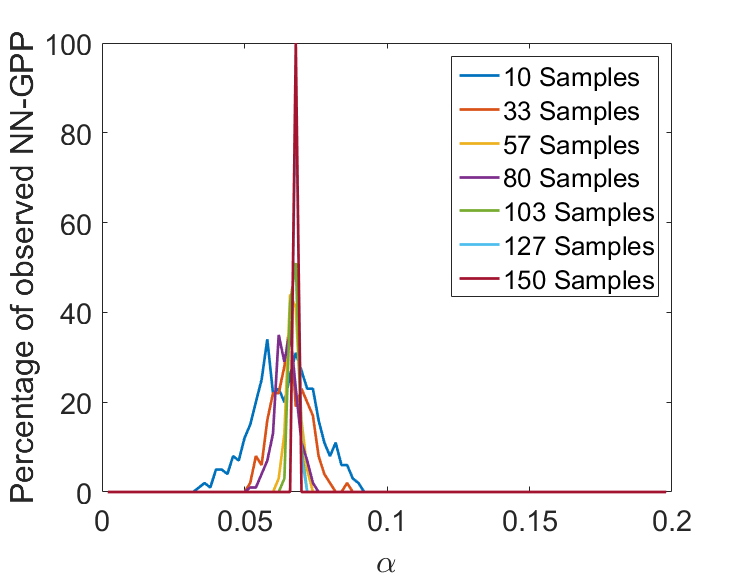}
		\caption{}
		\label{fig_dists_d11_g2}
	\end{subfigure}%
	
	\caption{(a) Function values and (b) directional derivatives along the full-batch steepest descent direction, $\boldsymbol{d}_n=-\nabla \mathcal{L}(\boldsymbol{x})$. The loss function is obtained from the Iris classification problem of Figure~\ref{fig_cfunc_full} \citep{Fisher1936}. The search direction is sampled by 100 points with sample sizes ranging from $|\mathcal{B}_{n,i}| = 10$ to $|\mathcal{B}_{n,i}| = M = 150$. This is repeated 100 times and the average number of minima and SNN-GPP found at every point is plotted. Minima are spread across the entire domain for most sample sizes in (a). Both minima and SNN-GPPs identify the true optimum, when the full batch is used. The spatial variance of SNN-GPP is bounded around the true optimum with increasing spread for decreasing sample size. However, even with the smallest batch size, $|\mathcal{B}_{n,i}|=10$, SNN-GPPs remain spatially bounded, unlike local minima, which are approximately uniformly spread along the sampled domain.}
	
	\label{fig_min_loc}
\end{figure}

Subsequently, we present empirical evidence that SNN-GPPs offer a more representative means of identifying optima in dynamic MBSS loss functions over local minima. Firstly, we consider the nature of function value and gradient information in practical neural network loss functions with dynamic MBSS. Figure~\ref{fig_min_loc} shows weights $x_8$ and $x_9$ of our Iris network, as these happen to have interesting curvature characteristics relative to one another. In Figure~\ref{fig_min_loc}(a) and (b) we show the full-batch loss and gradients in $x_8$ and $x_9$. As expected, these are smooth, continuous surfaces. As we introduce dynamic MBSS with a mini-batch size of $|\mathcal{B}_{n,i}|=10$, sampled uniformly with replacement, the shape characteristics of the loss function are lost to the variance in the discontinuities, see Figure~\ref{fig_min_loc}(a). Interestingly, the gradients are much less affected by discontinuities, particularly around the edges of the sampled domain. The centre of the domain still remains "noisy", but contrary to the function values, the shape characteristics of the dynamic MBSS gradients remain comparable to the full-batch equivalents.

Now consider a hypothetical LS-SGD update performed at iteration, $n$, where $\boldsymbol{d}_n= - \nabla \mathcal{L}(\boldsymbol{x})$, denotes the steepest descent direction of the full-batch loss function. We note the locations of all the minimizers and SNN-GPPs along $\boldsymbol{d}_n$ over 100 increments, $i$, of size $\alpha_{n,i} - \alpha_{n,i-1} = 0.002$. Local minima are identified where $\tilde{F}_n(\alpha_{n,i-1}) > \tilde{F}_n(\alpha_{min}) < \tilde{F}_n(\alpha_{n,i+1}) $ and SNN-GPPs where $ \tilde{F}_n'(\alpha_{n,i-1}) \leq 0$ and $\tilde{F}_n'(\alpha_{snnpgpp}) > 0$ . We repeat this procedure 100 times with different sample sizes $|\mathcal{B}_{n,i}|$ to approximate the likelihood of determining the locations of minima and SNN-GPPs in Figure~\ref{fig_min_loc}. The spatial distribution of local minima across the sampled domain approximate a uniform distribution. The location of the true optimum is identified by full batch $|\mathcal{B}_{n,i}| = M$. Conversely, the spatial variance of SNN-GPPs are constrained in what resembles a Binomial distribution around the true optimum, with variance inversely proportional to the sample size $|\mathcal{B}_{n,i}|$. The central message of these plots is that the spatial location of SNN-GPPs is bounded within $B_\epsilon$, making it a reliable metric to be implemented to determine step sizes in stochastic loss functions. 


\section{Algorithmic Details}
\label{sec_alg_deets}

We implement the Gradient-Only Line Search that is Inexact (GOLS-I) \citep{Kafka2019jogo}, which requires as inputs only a given descent direction $\boldsymbol{d}_n$ and an initial step size, $\alpha_{n,0}$, which is incrementally modified. We use increment counter, $i$, to determine the number of modifications made to $\alpha_{n,i}$ and corresponding function evaluations performed, until a final step size $\alpha_{n,I_n}$ is chosen. Thus, the total number of modifications in iteration, $n$ is $I_n$. Parameters which are set, but are open to modification by the user if desired are the step size scaling parameter $\eta \in \{\mathcal{R^+}\; |\; \eta > 1\}$ and the modified Wolfe condition parameter $c_2=0.9$. For the purpose of our discussion, we use the dynamic MBSS univariate $\tilde{F}_n(\alpha)$ and $\tilde{F}'_n(\alpha)$ notation for function evaluations relating to GOLS-I. However, note that GOLS-I can be used in the context of static and dynamic MBSS as well as full-batch sampling.

The method consists of two stages: 1) Determining the adequacy of the initial guess, and 2) searching for a sign change from $-$ to $+$ along a descent direction. For assessing the initial guess, a modified strong Wolfe condition is implemented, to give {\it initial accept condition} (IAC):
\begin{equation}
0 < \tilde{F}_n'(\alpha_{n,0}) \leq c_2 |\tilde{F}_n'(0)|,
\label{eq_IACondition}
\end{equation}
with $c_2 > 0$. If the initial guess satisfies Equation (\ref{eq_IACondition}), it is immediately accepted as $\alpha_{n,I_n}$ and no further modification is made to the step size for the current iteration. The IAC implies that initial step size has progressed over a sign change in the directional derivative in a controlled manner, whereby the magnitude of the directional derivative is still decreased. In practice, this condition was found to be superior to the standard strong Wolfe condition, $|F_n'(\alpha)| \leq c_2 |F_n'(0)|$ \citep{Arora2011}.

\begin{figure}[h!]
	\centering
	\includegraphics[width=0.6\linewidth]{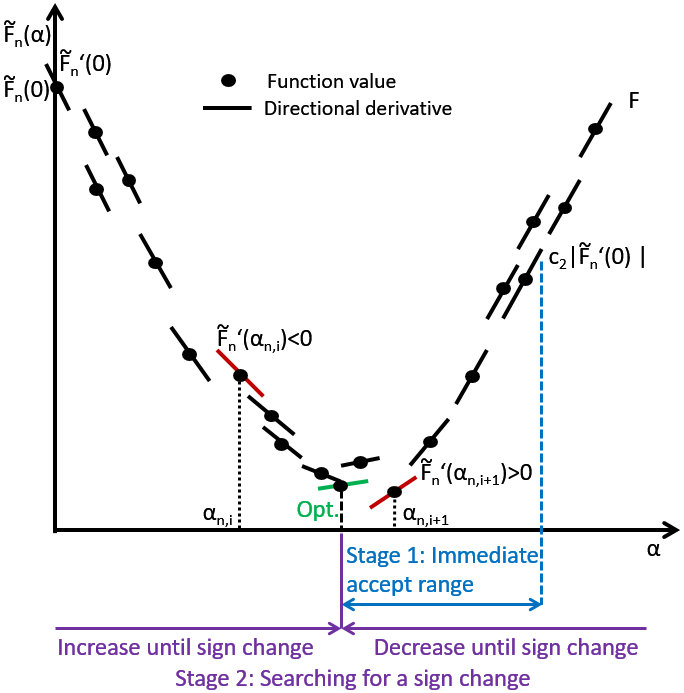}
	
	\caption{Schematic diagram of the Gradient-Only Line Search that is Inexact (GOLS-I): Dynamic MBSS results in discontinuous loss functions. $\tilde{F}'(\alpha_{n,0})$ is tested on the initial accept condition, Equation (\ref{eq_IACondition}), if this holds, $\alpha_{n,0}$ is accepted. Otherwise, the directional derivative sign of $\tilde{F}'(\alpha_{n,0})$ determines whether the step size needs to be increased by Equation (\ref{eq_alpha_increase}) or decreased by Equation (\ref{eq_alpha_decrease}) until an SNN-GPP is isolated.}
	\label{fig_diag_grad_inexact}
\end{figure}

If the initial accept condition is not satisfied, the algorithm enters stage 2, where the initial step size is increased or decreased by factor $\eta$ until a sign change is observed, i.e.

\begin{itemize}
	\item If $\tilde{F}_n'(\alpha_{n,0}) < 0,$ then 
	\begin{equation}
		\alpha_{n,i+1} = \eta \alpha_{n,i},
		\label{eq_alpha_increase}
	\end{equation}
	with $i := i + 1$ until $\tilde{F}_n'(\alpha_{n,i}) > 0$, or 
	\item if $\tilde{F}_n'(\alpha_{n,0}) > 0,$ then 
	\begin{equation}
		\alpha_{n,i+1} = \frac{\alpha_{n,i}}{\eta},
		\label{eq_alpha_decrease}
	\end{equation}
	with $i := i + 1$ until $\tilde{F}_n'(\alpha_{n,i}) < 0$.
\end{itemize}

This process is illustrated in Figure~\ref{fig_diag_grad_inexact}. An upper and lower limit is given for step sizes to ensure algorithmic stability in cases of monotonically decreasing or ascending search directions respectively. The maximum allowable step size is inspired by convergent fixed step sizes according to the Lipschitz condition \citep{Boyd2014}. We therefore choose the maximum step size conservatively as:
\begin{equation}
\alpha_{max} = \min(\frac{1}{\|\boldsymbol{d}_n\|_2}, 10^7).
\label{eq_max_step}
\end{equation}

Conservative updates are enforced by ensuring that $\alpha < \frac{1}{\|\boldsymbol{d}_n\|_2}$, which restricts the step size in the case of steep descent directions, but allows larger step sizes for more gradual descent directions. The absolute upper limit restricts overly large step sizes in the case of flat search directions. Taking the minimum of the two limits ensures that the line search can traverse both steep declines and flat planes, while remaining stable in the case of spurious mini-batch characteristics. 

The minimum step size avoids high computational cost, should the line search encounter an ascent direction. In such cases computational resources are wasted by continually decreasing the step size towards zero. Therefore the minimum step size is limited to:
\begin{equation}
\alpha_{min} = 10^{-8},
\label{eq_minalpha}
\end{equation} 
which in combination with the maximum step size results in an available step size range of 15 orders of magnitude. Unlike PrLS \citep{Mahsereci2017a}, GOLS-I does not limit the number of function evaluations per iteration, which makes the entire range of step size magnitudes available to the line search in pursuit of a directional derivative sign change from $-$ to $+$ along a descent direction.

In this study, the first iteration of GOLS-I ($n=0$), the initial guess is selected to be $\alpha_{0,0} = \alpha_{min}$. Therefore, GOLS-I is tasked with growing the step size from $\alpha_{min}$ to the desired magnitude as dictated by the presented loss function and descent direction. In subsequent iterations, the initial guess is set to be the final step size of the previous iteration, i.e. $\alpha_{n,0} = \alpha_{n-1,I_n}$. The pseudo code for GOLS-I is given in Algorithm \ref{alg_GOLSI}.

\begin{algorithm}[H]
	\DontPrintSemicolon 
	\KwIn{$\tilde{F}_n'(\alpha)$, $\boldsymbol{d}_n$ , $\alpha_{n,0}$}
	\KwOut{$\alpha_{n,I_n}$, $I_n$}
	
	Define constants: $\alpha_{min}=10^{-8}$, flag = 1, $\eta = 2$, $c_2=0.9$, $i=0$ \;
	$\alpha_{max} = min(\frac{1}{||\boldsymbol{d}_n ||_2}, 10^7)$ \;
	
	Evaluate $ \tilde{F}_n'(0) $, increment $i$ (or use saved gradient from last $ F_{n-1}'(\alpha_{n-1,I_{n-1}})$, to evaluate $ \tilde{\boldsymbol{g}}(\boldsymbol{x}_{n-1} + \alpha_{n-1,I_{n-1}} \cdot \boldsymbol{d}_{n-1})^T \boldsymbol{d}_n$ without incrementing $i$)\;
	
	\If{ $\alpha_{n,0} > \alpha_{max}$ }{
		$ \alpha_{n,0} = \alpha_{max} $ 
	}
	\If{ $\alpha_{n,0} < \alpha_{min}$}{
		$ \alpha_{n,0} = \alpha_{min} $ 
	}
	Evaluate $ \tilde{F}_n'(\alpha_{n,0})$, increment $i$ \;
	Define $tol_{dd} = |c_2 \tilde{F}_n'(0)|$ \;
	
	\If{ $\tilde{F}_n'(\alpha_{n,0}) > 0$ and $\alpha_{n,0} < \alpha_{max}$ }{
		flag = 1, decrease step size 
	}
	\If{ $\tilde{F}_n'(\alpha_{n,0}) < 0$ and $\alpha_{n,0} > \alpha_{min}$ }{
		flag = 2, increase step size  
	}
	\If{  $\tilde{F}_n'(\alpha_{n,0}) > 0$ and $\tilde{F}_n'(\alpha_{n,0}) < tol_{dd}$ }{
		flag = 0, immediate accept condition
	}
	
	\While{flag $>0$}{ 
		\If {flag = 2}{
			$\alpha_{n,i+1} = \alpha_{n,i} \cdot \eta$ \;
			Evaluate $ \tilde{F}_n'(\alpha_{n,i+1}) $ \;
			
			\If{ $\tilde{F}_n'(\alpha_{n,i+1}) \geq 0 $}{
				flag = 0 \; 
			}
			\If{ $\alpha_{n,i+1} > \frac{\alpha_{max}}{\eta}$}{
				flag = 0
			}	
		}
		\If {flag = 1}{
			$\alpha_{n,i+1} = \frac{\alpha_{n,i}}{\eta} $ \;
			Evaluate $ \tilde{F}_n'(\alpha_{n,i+1}) $ \;
			\If{$\tilde{F}_n'(\alpha_{n,i+1}) < 0$}{
				flag = 0
			}
			\If{ $\alpha_{n,i+1} < \alpha_{min} \cdot \eta$}{
				flag = 0
			}
		}
	}
	
	$\alpha_{n,I_n}$ = $\alpha_{n,i+1}$
	
	\caption{{\sc GOLS-I}: Gradient-Only Line Search that is Inexact}
	\label{alg_GOLSI}
\end{algorithm}

\subsection{Proof of Global Convergence for Full-Batch Sampling}

Concerning notation for the following proofs, the input variable $\boldsymbol{x}$ for loss functions is omitted in aid of brevity, such that $\mathcal{L}(\boldsymbol{x})$ is represented simply as $\mathcal{L}$. Therefore, suppose that the loss function $\mathcal{L}$ obtained from full batch sampling is smooth, coercive with a unique minimizer $\boldsymbol{x}^*$. Any Lipschitz function $\hat{\mathcal{L}}$ can be regularized to be coercive using Tikhonov regularization with a sufficient large regularization coefficient. 

The iteration updates of an optimization algorithm can be considered as a dynamical system in discrete time:
\begin{equation}
\boldsymbol{x}_{n+1} = \mathcal{D}(\boldsymbol{x}_{n}),\; \mathcal{D}:\mathcal{R}^p \rightarrow \mathcal{R}^p.
\end{equation}

It follows from Lyapunov's global stability theorem \citep{Lyapunov1992} in discrete time that any Lyapunov function $\Gamma(\boldsymbol{x})$ defined by positivity, coercive and strict decrease:
\begin{enumerate}
	\item Positivity: $\Gamma(\boldsymbol{0}) = \boldsymbol{0}$ and  $\Gamma(\boldsymbol{x}) > 0,\;\forall \boldsymbol{x}\neq\boldsymbol{0}$
	\item Coercive: $\Gamma(\boldsymbol{x}) \rightarrow \infty$ as $\boldsymbol{x}\rightarrow \infty$
	\item Strict descent: $\Gamma(\mathcal{D}(\boldsymbol{x}))  <  \Gamma(\boldsymbol{x}),\;\forall\;\boldsymbol{x} \neq \boldsymbol{0}$,
\end{enumerate}
results in $\boldsymbol{x}_{n}\rightarrow \boldsymbol{0}$ as $n \rightarrow \infty,\;\forall \;\boldsymbol{x}_0 \in \mathcal{R}^p$.

\begin{theorem}
	Let $f(\boldsymbol{x})$ be any smooth coercive function with a unique global minimum $\boldsymbol{x}^*$, for $\boldsymbol{x}_{n+1} = \mathcal{D}(\boldsymbol{x}_n),\;\forall\;\boldsymbol{x}_n \neq \boldsymbol{x}_n^*$ restricted such that $f(\beta \boldsymbol{x}_{n+1} + (1-\beta) \boldsymbol{x}_n)  <  f(\boldsymbol{x}_n),\;\forall\;\beta \in (0,1]$. Then $ \mathcal{D}$ will result in updates that are globally convergent.
\end{theorem}

Let the error at iteration $n$ be given by $\boldsymbol{e}_n := \boldsymbol{x}_n - \boldsymbol{x}^*$ for which we can construct the Lyapunov function $\Gamma(\boldsymbol{e}) = f(\boldsymbol{e} + \boldsymbol{x}^*) - f(\boldsymbol{x}^*)$. It follows that $\Gamma(\boldsymbol{0}) = \boldsymbol{0}$ and that  $\Gamma(\boldsymbol{e}) > 0,\;\forall\;\boldsymbol{e}\neq\boldsymbol{0}$, since $\boldsymbol{x}^*$ is a unique global minimum of $f$. 


At every iteration our line search update locates an SNN-GPP along the descent direction $\boldsymbol{d}_n$, by locating a sign change from negative to positive along $\boldsymbol{d}_n$. \cite{Wilke2013} proved this to be equivalent to minimizing along $\boldsymbol{d}_{n}$ when $f(\boldsymbol{x}_n + \alpha\boldsymbol{d}_n)$ is smooth and the sign of the directional derivative 
$\nabla^\textrm{T} f(\boldsymbol{x}_n + \alpha\boldsymbol{d}_n) \boldsymbol{d}_n$,
is negative $\forall\;\alpha \in [0,\alpha^*_n)$ along $\boldsymbol{d}_{n}$. Here, $\alpha^*_n$ defines the step size to the first optimum along the search direction $\boldsymbol{d}_{n}$. It is therefore guaranteed that $f(\boldsymbol{x}_{n+1}) <  f(\boldsymbol{x}_n)$ at every iteration $n$. In addition, $f(\beta \boldsymbol{x}_{n+1} + (1-\beta) \boldsymbol{x}_n)  <  f(\boldsymbol{x}_n),\;\forall\;\beta \in (0,1]$ ensures that for our choice of discrete dynamical update $\mathcal{D}$, we can always make progress unless $\boldsymbol{x}_n = \boldsymbol{x}^*$. Hence,  for any $\boldsymbol{e}_n \neq \boldsymbol{0}$ it follows that 
$$\Gamma(\boldsymbol{e}_{n+1}) = f(\boldsymbol{x}_{n+1} - \boldsymbol{x}^* + \boldsymbol{x}^*) - f(\boldsymbol{x}^*) < 
f(\boldsymbol{x}_{n} - \boldsymbol{x}^* + \boldsymbol{x}^*) - f(\boldsymbol{x}^*) = \Gamma(\boldsymbol{e}_{n}).$$ 

It then follows from Lyaponov's global stability theorem that $\boldsymbol{e}_n\rightarrow \boldsymbol{0}$ as $n\rightarrow \infty.$ Hence $\forall\;\boldsymbol{x}_0$ we have that $\boldsymbol{x}_n\rightarrow \boldsymbol{x}^*$, which proves that finding an SNN-GPP at every iteration $n$ results in a globally convergent strategy.

\subsection{Proof of Global Convergence for Dynamic Mini-Batch Sub-Sampling}

Consider the discontinuous loss function $\tilde{L}$ obtained through dynamic mini-batch sub-sampling with smooth expected response $E[\tilde{L}]$ and unique expected minimizer $\boldsymbol{x}^*$. Assume that the function $\tilde{L}$ is directional derivative coercive (see \cite{Wilke2013}) around a ball $\boldsymbol{x}\in\hat{B}_\epsilon (\boldsymbol{x}) =  \{\boldsymbol{q} \;|\; \|\boldsymbol{q} - \boldsymbol{x}^*\|	 \geq \epsilon  \}$  of given radius $\epsilon \in \mathcal{R} > 0$ that is centred around the expected minimizer $\boldsymbol{x}^*$. This implies that for given radius $\epsilon$ and for any point outside the ball $\boldsymbol{x}_1 \in \hat{B}_\epsilon (\boldsymbol{x}_1)$ and any point inside the ball $\boldsymbol{x}_2\in {B}_\epsilon (\boldsymbol{x}_2) =  \{\boldsymbol{q} \;|\; \|\boldsymbol{q} - \boldsymbol{x}^*\| < \epsilon  \}$ with $\boldsymbol{u} =  \boldsymbol{x}_1 - \boldsymbol{x}_2 $ the following must hold:
\begin{equation}
\nabla f(\boldsymbol{x}_1)^T \boldsymbol{u} > 0.
\end{equation}
As before, the iteration updates of an optimization algorithm can be considered as a dynamical system in discrete time:
\begin{equation}
\boldsymbol{x}_{n+1} = \mathcal{D}(\boldsymbol{x}_{n}),\; \mathcal{D}:\mathcal{R}^p \rightarrow \mathcal{R}^p.
\end{equation}

We relax Lyapunov's global stability theorem in discrete time for dynamic MBSS discontinuous functions, such that any smooth expected Lyapunov function $E[\Gamma(\boldsymbol{x})]$ defined by expected positivity, coercive and expected strict decrease around a ball $\boldsymbol{x}\in\hat{B}_\epsilon (\boldsymbol{x}) =  \{\boldsymbol{q} \;|\; \|\boldsymbol{q} - \boldsymbol{x}^*\| \geq \epsilon  \}$   of given radius $\epsilon \in \mathcal{R} > 0$, with:
\begin{enumerate}
	\item Expected positivity: $E[\Gamma(\boldsymbol{0})] = \boldsymbol{0}$ and  $E[\Gamma(\boldsymbol{x})] > 0,\;\forall \; \boldsymbol{x} \neq  \boldsymbol{0}$
	\item Coercive: $\Gamma(\boldsymbol{x}) \rightarrow \infty$ as $\boldsymbol{x}\rightarrow \infty$
	\item Directional derivative coercive for any point  $\boldsymbol{x} \in \hat{B}_\epsilon (\boldsymbol{x}) =  \{\boldsymbol{q} \;|\; \|\boldsymbol{q} - \boldsymbol{x}^*\|	 > \epsilon  \}$ of radius $\epsilon \in \mathcal{R} > 0$
	\item Expected strict descent: $E[\Gamma(\mathcal{D}(\boldsymbol{x}))]  <  E[\Gamma(\boldsymbol{x})],\;\forall\;\boldsymbol{x} \neq \boldsymbol{0}$,
\end{enumerate}
results in $\boldsymbol{x}_{n} \in {B}_\epsilon (\boldsymbol{x}_n) =  \{\boldsymbol{q} \;|\; \|\boldsymbol{q} - \boldsymbol{x}^*\|	< \epsilon  \}$ as $n \rightarrow \infty,\;\forall \;\boldsymbol{x}_0 \in \mathcal{R}^p$.

\begin{theorem}
	Let $f(\boldsymbol{x})$ be any smooth expected coercive function with a unique expected global minimum $\boldsymbol{x}^*$ that is directional derivative coercive around a ball $\hat{B}_\epsilon (\boldsymbol{x}) =  \{\boldsymbol{q} \;|\; \|\boldsymbol{q} - \boldsymbol{x}^*\| \geq \epsilon \}$ of radius $\epsilon \in \mathcal{R} > 0$. Then $\boldsymbol{x}_{n+1} = \mathcal{D}(\boldsymbol{x}_n),\;\forall\;\boldsymbol{x}_n \in \hat{B}_\epsilon (\boldsymbol{x}_n)$ is restricted such that $\nabla^\textrm{T} f(\boldsymbol{x}_n + \alpha\boldsymbol{d}_n) \boldsymbol{d}_n < 0,\;\forall\;\alpha \in [0,\alpha^*_n)$ along descent direction $\boldsymbol{d}_{n}$. Then $ \mathcal{D}$ will result in updates that globally converge to the ball $B_\epsilon(\boldsymbol{x}) = \{\boldsymbol{q} \;|\; \|\boldsymbol{q} - \boldsymbol{x}^*\|	 < \epsilon  \}$ of radius $\epsilon \in \mathcal{R} > 0$ centered around $ \boldsymbol{x}^*$.
\end{theorem}

Let the error at iteration $n$ be given by $\boldsymbol{e}_n := \boldsymbol{x}_n - \boldsymbol{x}^*$ for which we can construct the Lyapunov function $\Gamma(\boldsymbol{e}) = f(\boldsymbol{e} + \boldsymbol{x}^*) - f(\boldsymbol{x}^*)$ and expected Lyapunov function $E[\Gamma(\boldsymbol{e})] = E[f(\boldsymbol{e} + \boldsymbol{x}^*)] - E[f(\boldsymbol{x}^*)]$. It follows that $E[\Gamma(\boldsymbol{0})] = \boldsymbol{0}$ and that $E[\Gamma(\boldsymbol{e})] > 0,\;\forall\;\boldsymbol{e}\neq\boldsymbol{0}$, since $\boldsymbol{x}^*$ is a unique expected global minimum of $f$.

At every iteration our line search update locates an SNN-GPP along the descent direction $\boldsymbol{d}_n$, by locating a sign change from negative to positive along $\boldsymbol{d}_n$. 
Since the function is smooth expected coercive and directional derivative coercive around a ball 
$\hat{B}_\epsilon (\boldsymbol{x})$, expected descent follows $E[\Gamma(\mathcal{D}(\boldsymbol{x}))]  <  E[\Gamma(\boldsymbol{x})],\;\forall\;\boldsymbol{x} \notin B_\epsilon(\boldsymbol{x}) = \{\boldsymbol{q} \;|\; \|\boldsymbol{q} - \boldsymbol{x}^*\|	 < \epsilon  \}$ of radius $\epsilon \in \mathcal{R} > 0$.
It is therefore guaranteed that $E[f(\boldsymbol{x}_{n+1})] <  E[f(\boldsymbol{x}_n)]$ at every iteration $n$. In addition, $E[f(\beta \boldsymbol{x}_{n+1} + (1-\beta) \boldsymbol{x}_n)]  <  E[f(\boldsymbol{x}_n)],\;\forall\;\beta \in (0,1]$ ensures that for our choice of discrete dynamical update $\mathcal{D}$, we can always make progress unless $\boldsymbol{x}_n \in B_\epsilon(\boldsymbol{x}) = \{\boldsymbol{q} \;|\; \|\boldsymbol{q} - \boldsymbol{x}^*\| < \epsilon  \}$. In addition, since the function is directional derivative coercive around the ball ${B}_\epsilon (\boldsymbol{x}_n)$, any point $\boldsymbol{x} \in {B}_\epsilon (\boldsymbol{x}_n)$ remains in ${B}_\epsilon (\boldsymbol{x}_n)$ due to the update requirement of a sign change from negative to positive along the descent direction. Hence,  for any $\boldsymbol{e}_n$ such that $\|\boldsymbol{e}_n\| > \epsilon$ it follows that: 
$$E[\Gamma(\boldsymbol{e}_{n+1})] = E[f(\boldsymbol{x}_{n+1} - \boldsymbol{x}^* + \boldsymbol{x}^*)] - E[f(\boldsymbol{x}^*)] < 
E[f(\boldsymbol{x}_{n} - \boldsymbol{x}^* + \boldsymbol{x}^*)] - E[f(\boldsymbol{x}^*)] = E[\Gamma(\boldsymbol{e}_{n})].$$ 

It then follows from Lyuaponov's relaxed global stability theorem that $\boldsymbol{e}_n \in B_\epsilon(\boldsymbol{x})$ as $n\rightarrow \infty.$ Hence $\forall\;\boldsymbol{x}_0$ we have that $\boldsymbol{x}_n \in \hat{B}_\epsilon(\boldsymbol{x}_n) \rightarrow \boldsymbol{x}_n \in B_\epsilon(\boldsymbol{x}_n)$  as $n\rightarrow\infty$, which proves that finding an SNN-GPP at every iteration $n$ results in a globally converges to the ball ${B}_\epsilon(\boldsymbol{x}_n)$.

\section{Numerical Studies}
\label{sec_numStudies}

The network architectures and problems implemented in the numerical investigations of this study are predominantly taken from the work of \citep{Mahsereci2017a}, which serves as the benchmark against which we compare GOLS-I. The work by \cite{Mahsereci2017a} introduces PrLS, which is used to estimate step sizes in stochastic gradient descent \citep{Robbins1951} with a line search (LS-SGD) and compared to standard SGD with {\it a priori} fixed step sizes. We directly compare the performances of PrLS and GOLS-I for determining step sizes for LS-SGD using the Matlab code for PrLS supplied\footnote{https://ei.is.tuebingen.mpg.de/publications/mahhen2015} in the relevant article \citep{Mahsereci2017a}. The datasets we consider are:

\begin{itemize}
	\item Breast Cancer Wisconsin Diagnostic (BCWD) Dataset \citep{Wolberg1993}, a binary classification problem, distinguishing between "benign" and "malignant" tumors, using 30 different features;
	\item MNIST Dataset \citep{Lecun1998}, a multi-class classification problem with images of handwritten digits from 0 to 9 in grey-scale with a resolution of 28x28 pixels; and
	\item CIFAR10 \citep{Krizhevsky2009}, a multi-class classification problem with images of 10 natural objects such as deer, cats, dogs, ships, etc. with colour images of resolution 32x32.
\end{itemize}

Table~\ref{tbl_datasets} gives further details about the datasets, as well as parameters specific to their implementations such as: The neural network architectures used, number of function evaluations to which training is limited and mini-batch sizes used. These details are implemented as  prescribed by \citet{Mahsereci2017a}. GOLS-I was implemented in both Matlab and PyTorch 1.0. The Matlab implementations are used for direct comparisons between GOLS-I and PrLS, while the PyTorch implementation demonstrates how GOLS-I compares to fixed step sizes. This latter implementation also demonstrates the ease by which GOLS-I is transferred to GPU-capable computing platforms. All datasets were pre-processed using the standard transformation (Z-score) for each input dimension in the dataset.

\begin{table}[h]
	\centering
	\scalebox{0.99}{
		\begin{tabularx}{\textwidth}{|p{14mm}|p{15mm}|p{9mm}|p{10mm}|p{13mm}|p{28mm}|p{10mm}|X|}
			\hline 
			\textbf{Datset} & \textbf{Training obs.} & \textbf{Test obs.} & \textbf{Input dim.} & \textbf{Output dim.} & \textbf{Net structure} & \textbf{Max. F.E.} & $ |\mathcal{B}_{n,i}| $ \textbf{in}  \newline \textbf{training }\\ 
			\hline 
			\textbf{BCWD} & 400 & 169 & 30 & 2 & Log. Regression, NetPI, NetPII  & 3000/ 100000 & 10,50, 100,400   \\ 
			\hline 
			\textbf{MNIST} & 50000 & 10000 & 784 & 10 & NetI, NetII & 40000 &  10,100, 200 \\ 
			\hline 
			\textbf{CIFAR} & 10000 (Batch1) & 10000  & 3072 & 10 & NetI, NetII & 10000 &  10,100, 200  \\ 
			\hline 
		\end{tabularx} 
	}
	\caption{Relevant parameters related to the datasets used in numerical experiments. }
	\label{tbl_datasets}
\end{table}

Following \cite{Mahsereci2017a}, both MNIST and CIFAR10 are implemented using two different network architectures, NetI and NetII, while the smaller BCWD dataset was trained using logistic regression and two single hidden layer fully connected networks, NetPI and NetPII. The latter two network architectures are borrowed from the work of \cite{Prechelt1994}, in which a modified version of the BCWD dataset was implemented. The parameters governing the different implementations are summarized in Table~\ref{tbl_nets}. This constitutes a total of 7 combinations of different datasets, architectures and loss functions used in the numerical study. All networks are fully connected, and although the detail given concerning the hidden layers of the network excludes the biases in Table~\ref{tbl_nets}, they are included in the implementations. \cite{Mahsereci2017a} have stated that a normal distribution was used to initialize all networks. However, we found that the problems using NetII would not converge using normally distributed weight initializations unless the variance was reduced to 0.1 for MNIST and 0.01 for CIFAR10 respectively. The latter variance scaling was also adopted for the NetI implementations of CIFAR10. For each problem, 10 training runs were conducted, where training and test classification errors evaluated, as well as estimated step sizes are noted during training. The resolution of the classification error plots in the log domain is limited by the size of the respective training and test datasets. For the BCWD dataset, this is the full test dataset, while for MNIST and CIFAR10, the classification errors are evaluated for a random subsample of 1000 observations from the training and test datasets respectively. Therefore, to avoid bad scaling in error plots for cases where the classification error is 0, a numerical constant of $1\cdot 10^{-4}$ was added to classification error calculations. Therefore the value $1\cdot 10^{-4}$ on classification error plots represents absolute zero.

For our PyTorch implementation of the WDBC logistic regression problem, we selected three constant step sizes, each one order of magnitude apart, $\alpha_{n,I_n} \in \{1, 10, 100 \}$, ensuring that the full training performance modality is captured. This means that the small fixed step size represents a slow and overly conservative learning rate that leads to wasted gradient computations during training. The medium fixed step results in an effective and efficient learning rate with desired convergence performance and the large fixed step in training that is too aggressive and usually leads to detrimental performance. GOLS-I's training performance is compared to the three constant learning rates, to investigate its feasibility in replacing {\it a priori} determined fixed step sizes.

\begin{table}[h]
	\centering
	\scalebox{0.99}{
		\begin{tabularx}{\textwidth}{|p{20mm}|p{26mm}|p{21mm}|p{33mm}|X|}
			\hline 
			\textbf{Network} & \textbf{Hidden layer structure} & \textbf{Activation function} & \textbf{Initialization} & \textbf{Loss function} \\ 
			\hline 
			\textbf{Logistic regression} & N/A & Sigmoid & $\mathcal{N}(0,I)$ & BCE \\ 
			\hline 
			\textbf{NetPI} & 32 & Sigmoid & $\mathcal{N}(0,I)$ & BCE \\ 
			\hline 
			\textbf{NetPII} & 32 & Sigmoid & $\mathcal{N}(0,I)$ & MSE \\ 
			\hline 
			\textbf{NetI} & 800 & Sigmoid & $\mathcal{N}(0,I)$ MNIST $\mathcal{N}(0,0.01I)$ CIFAR & BCE \\ 
			\hline 
			\textbf{NetII} & 1000,500,250 & Tanh & $\mathcal{N}(0,0.1I)$ MNIST $\mathcal{N}(0,0.01I)$ CIFAR & MSE \\ 
			\hline 
		\end{tabularx} 
	}
	\caption{Parameters and settings governing the implemented network architectures and their training.}
	\label{tbl_nets}	
\end{table}

There is a significant difference in the information required by PrLS and GOLS-I. Apart from using function value and gradient information, which is readily available in most neural network software, such as Tensorflow \citep{Tensorflow2} and PyTorch \citep{PyTorch1}, PrLS also requires variance estimates of both function values and gradients, which are not standard outputs. Conversely, GOLS-I requires only gradients to be evaluated, in order to calculate directional derivatives. However, since gradients are evaluated via backpropagation, function values are evaluated as a by-product. In order to be consistent with the information presented to PrLS and GOLS-I in our direct comparisons (using Matlab), we consider a function evaluation to be: A function value, the respective gradient, as well as the in-batch variance estimates of function value and gradients as prescribed by \cite{Mahsereci2017a}. Of this information, GOLS-I then only uses the gradient vector, while PrLS uses all of the above. In our PyTorch implementation of GOLS-I, only the gradient vector is evaluated. 

Both PrLS and GOLS-I require an initial guess at the beginning of a training run, after which the step size of the previous iteration is used as the initial guess for the next iteration, i.e. $\alpha_{n,0}=\alpha_{n-1,I_n}$. The starting initial guess for PrLS is set as $\alpha_{0,0} = 10^{-4}$, as prescribed by \cite{Mahsereci2017a}, while GOLS-I uses a more conservative $\alpha_{0,0} = 10^{-8}$, as discussed in Section \ref{sec_alg_deets}. The minimum number of function evaluations per iteration is 1 for both PrLS and GOLS-I, when the initial condition is satisfied. Conversely, the maximum number of function evaluations for PrLS is capped at 7, while GOLS-I is uncapped in function evaluations, but practically capped by the range of available step sizes until $\alpha_{max}$ is reached.

\subsection{Software}
In order to aid transparency and reproducibility, we have made our code available at \url{https://github.com/gorglab/GOLS}. The given repositories include accessible adaptations of the source code used in our investigations. We also include examples of other GOLS methods developed in \cite{Kafka2019jogo}, which are GPU compatible. The examples presented are self contained to create an environment suited to exploring GOLS.

\section{Results}

Results are categorized according to the datasets and respective architectures investigated, where the training and test classification errors, as well as estimated step sizes are shown for different mini-batch sizes, $|\mathcal{B}_{n,i}|$. Note, that classification error plots are shown in terms of function evaluations, while step sizes are shown in terms of iterations. This allows the training performances between PrLS and GOLS-I to be compared in terms of computational cost, while also showing the difference in function evaluations per iteration in step size plots. 

\subsection{The Breast Cancer Wisconsin Diagnostic (BCWD) Dataset with Logistic Regression, NetPI and NetPII}
\label{sec_BCWD}

\begin{figure}[h]
	\centering
	
	\begin{subfigure}{.245\textwidth}
		\centering 
		\includegraphics[width=0.99\linewidth]{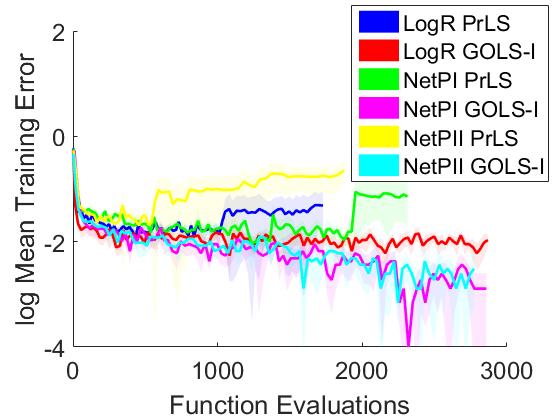}
	\end{subfigure}%
	\begin{subfigure}{.245\textwidth}
		\centering
		\includegraphics[width=0.99\linewidth]{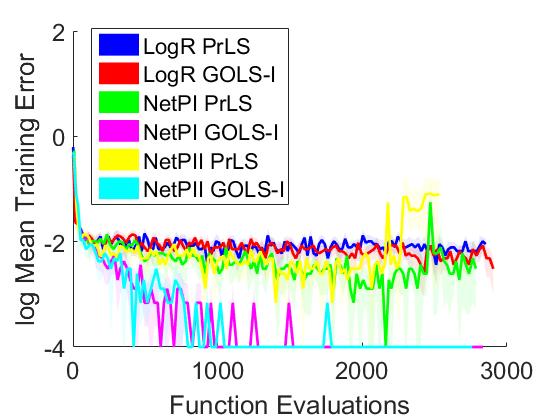}
	\end{subfigure}%
	\begin{subfigure}{.245\textwidth}
		\centering 
		\includegraphics[width=0.99\linewidth]{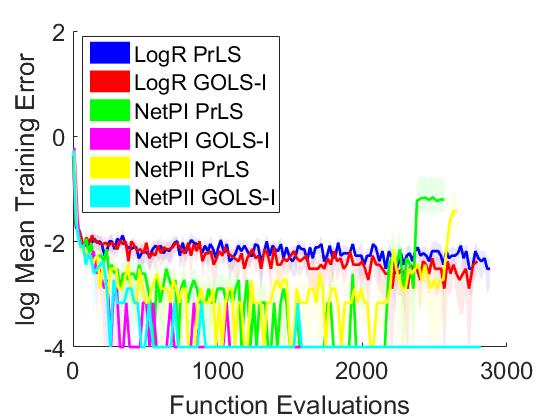}
	\end{subfigure}%
	\begin{subfigure}{.245\textwidth}
		\centering 
		\includegraphics[width=0.99\linewidth]{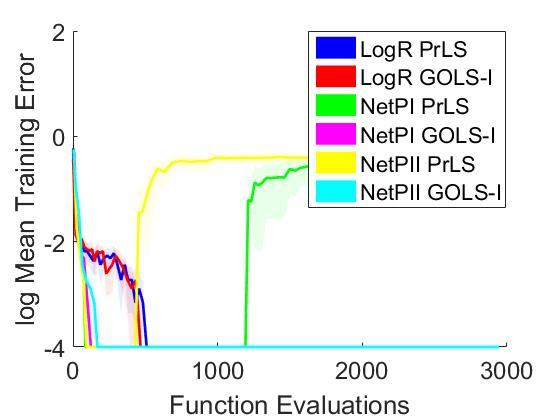}
	\end{subfigure}%
	
	\begin{subfigure}{.245\textwidth}
		\centering 
		\includegraphics[width=0.99\linewidth]{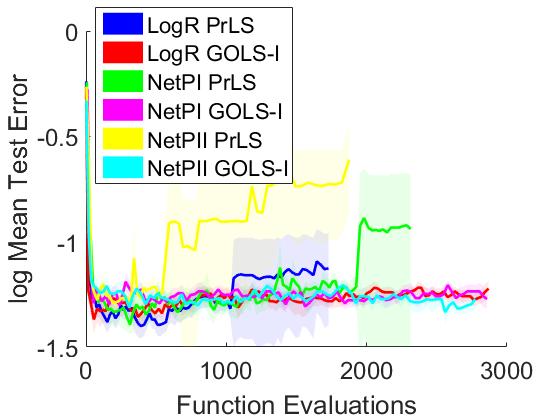}
	\end{subfigure}%
	\begin{subfigure}{.245\textwidth}
		\centering
		\includegraphics[width=0.99\linewidth]{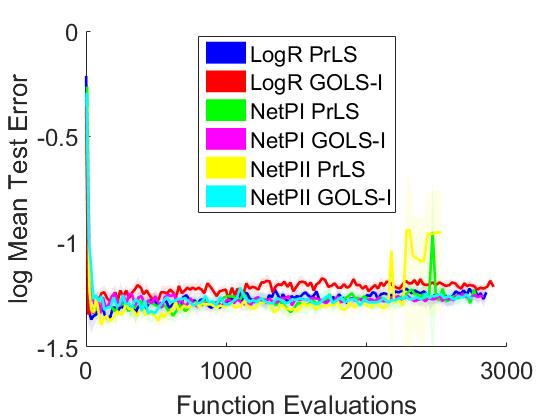}
	\end{subfigure}%
	\begin{subfigure}{.245\textwidth}
		\centering 
		\includegraphics[width=0.99\linewidth]{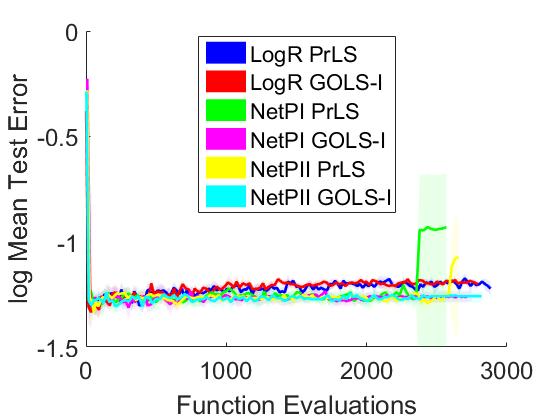}
	\end{subfigure}%
	\begin{subfigure}{.245\textwidth}
		\centering 
		\includegraphics[width=0.99\linewidth]{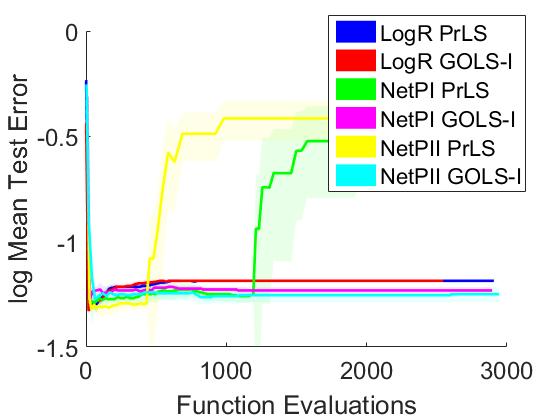}
	\end{subfigure}%
	
	\begin{subfigure}{.245\textwidth}
		\centering 
		\includegraphics[width=0.99\linewidth]{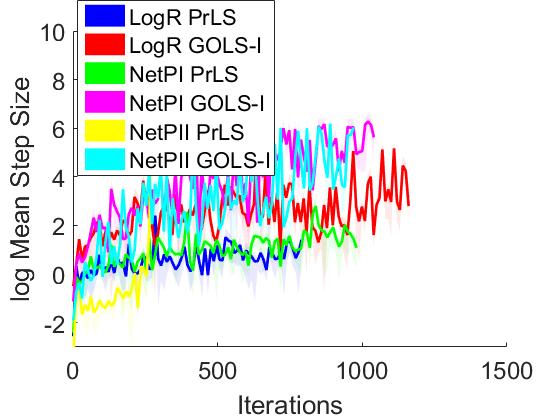}
		\caption{$|\mathcal{B}_{n,i}| = 10$}
	\end{subfigure}%
	\begin{subfigure}{.245\textwidth}
		\centering
		\includegraphics[width=0.99\linewidth]{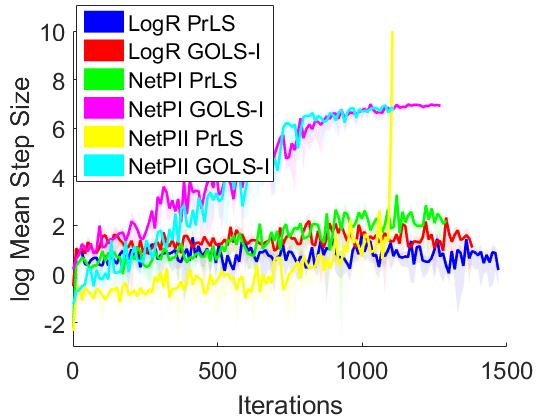}
		\caption{$|\mathcal{B}_{n,i}| = 50$}
	\end{subfigure}%
	\begin{subfigure}{.245\textwidth}
		\centering 
		\includegraphics[width=0.99\linewidth]{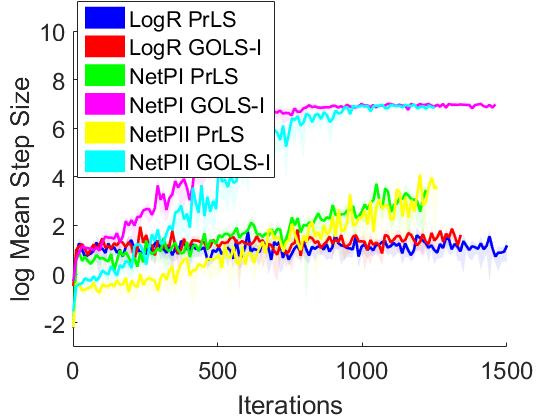}
		\caption{$|\mathcal{B}_{n,i}| = 100$}
	\end{subfigure}%
	\begin{subfigure}{.245\textwidth}
		\centering 
		\includegraphics[width=0.99\linewidth]{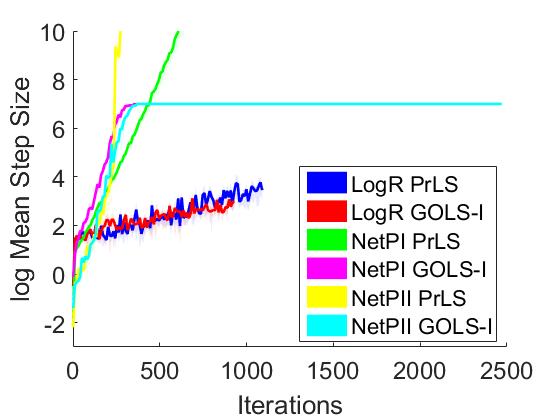}
		\caption{$M = 400$}
	\end{subfigure}%
	
	\caption{Log training error, log training loss, log test error and the log of the step sizes as obtained with various batch sizes for the BCWD dataset problem.}
	\label{fig_cancer_comp}
\end{figure}

Training and test classification error as well as corresponding step sizes for the different network architectures used with the BCWD dataset are shown in Figure~\ref{fig_cancer_comp}. Note, that in order to clearly distinguish and compare the performances of the investigated architectures for this problem, we plot the mean over 10 runs using a solid line, while the variance cloud is indicated by a shaded area around the mean. This representation is not used for subsequent figures, where each of the 10 runs is plotted individually, as determined by \cite{Mahsereci2017a}.

In the logistic regression (LogR) example, the performance of GOLS-I and PrLS is similar for mini-batch sizes of $|\mathcal{B}_{n,i}| \geq 50 $. In these cases, the step sizes estimated by both line searches are also of comparable magnitude. This shows that both PrLS and GOLS-I automatically adapt their starting initial guesses from $\alpha_{0,0}=1\cdot 10^{-4}$ and $\alpha_{0,0}=1\cdot 10^{-8}$ respectively to magnitudes around 1 and larger.  For the logistic regression problem we observe an increase in step sizes as a function of iterations for both methods. On average, the number of gradient evaluations per iteration is in the low 2's for the BCWD dataset, but can be up to 17 for individual iterations of GOLS-I. However, on average we see that GOLS-I and PrLS also complete a similar number of iterations for a fixed maximum number of function evaluations. This suggests, that the performance between the two line searches is competitive, an assertion supported by the almost indistinguishable training, test and step size plots for the full batch case, $M=400 $.

However, for $|\mathcal{B}_{n,i}| = 10 $ PrLS exhibits unstable behaviour, encountering numerical difficulties that resulted in early termination. This is in contrast to the results shown by \cite{Mahsereci2017a}, which suggest stability with $|\mathcal{B}_{n,i}| = 10 $ for 10,000 function evaluations. We observe this unstable behaviour for all three architectures trained with PrLS and $|\mathcal{B}_{n,i}| = 10 $, while conversely, GOLS-I completed all training runs with $|\mathcal{B}_{n,i}| = 10 $.

Next, consider the results of NetPI and NetPII with $|\mathcal{B}_{n,i}| \geq 50 $. Training performance of both line searches improves as a function of increasing mini-batch size. However, the performance difference between GOLS-I and PrLS becomes more distinct for these network architectures. The use of GOLS-I produces instances where the training classification errors are $0$. In such cases, the lower bound of the y-axis is limited to the numerical nugget of $10^{-4}$. The smallest non-zero error for BCWD is $log(\frac{1}{400}) \approx -2.6$. For GOLS-I with $|\mathcal{B}_{n,i}| = 50 $, most training runs reach zero training classification error after approximately 1000 function evaluations. This is accelerated to 500-800 function evaluations for $|\mathcal{B}_{n,i}| = 100 $. The use of PrLS does not produce perfect training results for $|\mathcal{B}_{n,i}| = 50 $, but does improve for $|\mathcal{B}_{n,i}| = 100 $, first reaching $10^{-4}$ after $\pm 1200$ function evaluations. This difference in performance is due to the difference in step sizes determined by GOLS-I and PrLS for $|\mathcal{B}_{n,i}| \in \{50,100\}$. GOLS-I grows its step size at a significantly faster rate than PrLS for both NetPI and NetPII, reaching $\alpha_{max}$ while still remaining stable. This means that the step sizes determined by PrLS for $|\mathcal{B}_{n,i}| \in \{50,100\}$ are more conservative. The step size behaviour of PrLS changes significantly for $M = 400$, where the step sizes of both methods grow at a similar rate until $\pm 400$ function evaluations, after which GOLS-I reaches its maximum step size, but PrLS continues to grow its steps. This has the consequence that PrLS becomes unstable after latest 1200 function evaluations. We postulate, that this weakness of PrLS is related to cases where the norm of the gradient vector approaches 0 towards the end of training, an occurrence which we observe in training and test errors for all mini-batch sizes.

Apart from the late-training divergence of PrLS, the test classification errors between line search methods and network architectures are largely the same for this problem. Overfitting occurs readily after around 200 function evaluations, with the logistic regression implementations showing higher test classification errors in later training than both NetPI and NetPII. This indicates, that both PrLS and GOLS-I are in principle capable of fitting the given architectures to the BCWD dataset. However, for this example GOLS-I has improved training performance to PrLS, as well as exhibiting increased algorithmic stability.

\begin{figure}[h]
	\centering
	
	\begin{subfigure}{0.245\textwidth}
		\centering 
		\includegraphics[width=0.99\linewidth]{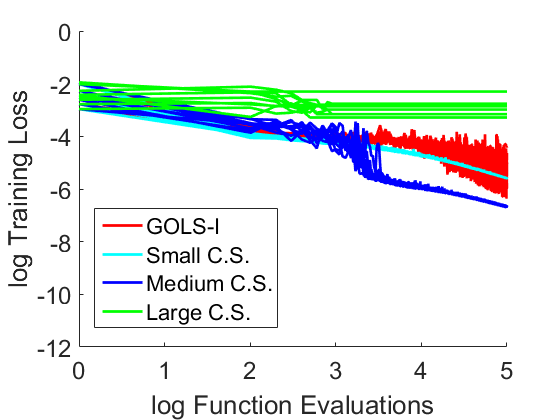}
	\end{subfigure}%
	\begin{subfigure}{0.245\textwidth}
		\centering
		\includegraphics[width=0.99\linewidth]{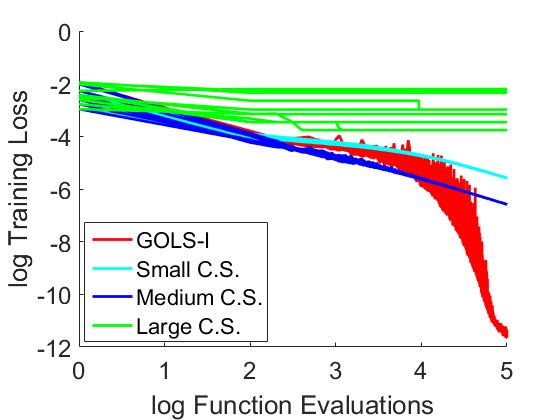}
	\end{subfigure}%
	\begin{subfigure}{0.245\textwidth}
		\centering 
		\includegraphics[width=0.99\linewidth]{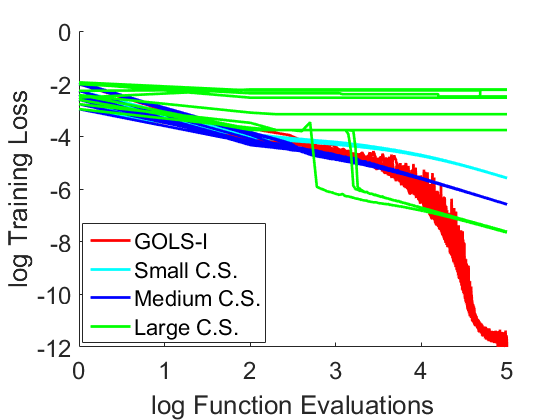}
	\end{subfigure}%
	\begin{subfigure}{0.245\textwidth}
		\centering 
		\includegraphics[width=0.99\linewidth]{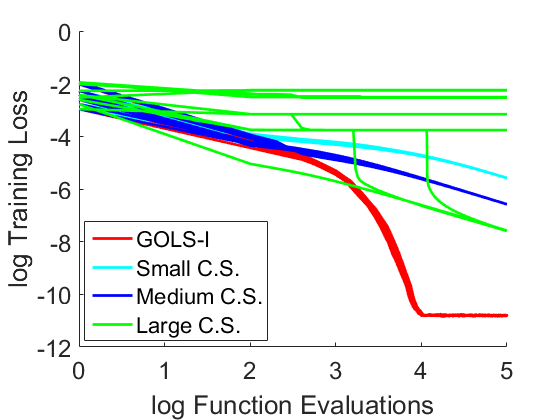}
	\end{subfigure}%
	
	\begin{subfigure}{0.245\textwidth}
		\centering 
		\includegraphics[width=0.99\linewidth]{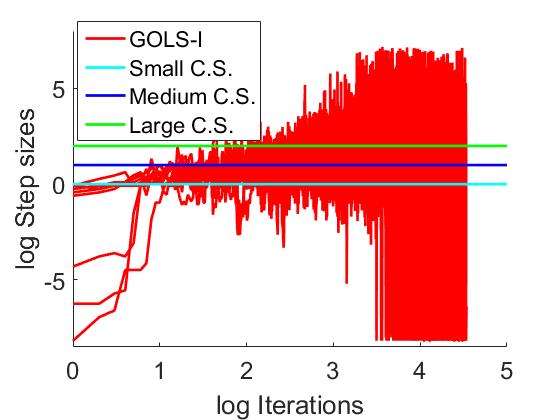}
		\caption{$|\mathcal{B}_{n,i}| = 10$}
	\end{subfigure}%
	\begin{subfigure}{0.245\textwidth}
		\centering
		\includegraphics[width=0.99\linewidth]{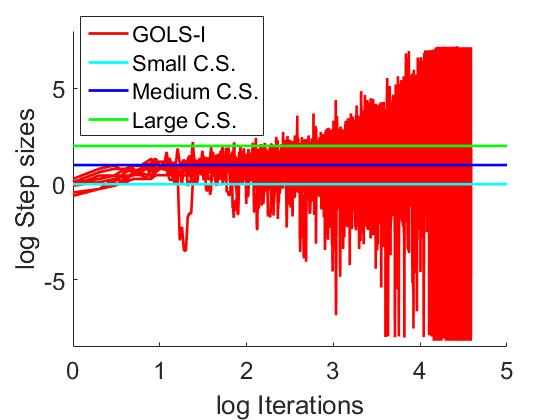}
		\caption{$|\mathcal{B}_{n,i}| = 50$}
	\end{subfigure}%
	\begin{subfigure}{0.245\textwidth}
		\centering 
		\includegraphics[width=0.99\linewidth]{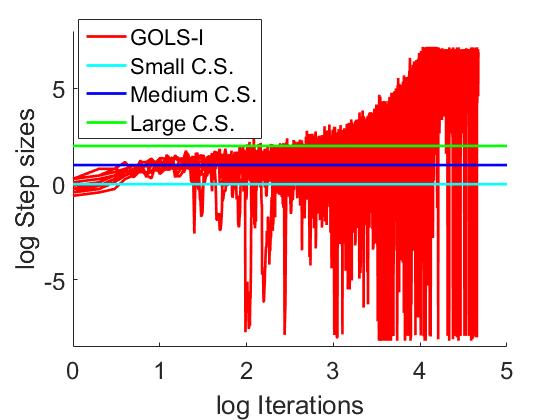}
		\caption{$|\mathcal{B}_{n,i}| = 100$}
	\end{subfigure}%
	\begin{subfigure}{0.245\textwidth}
		\centering 
		\includegraphics[width=0.99\linewidth]{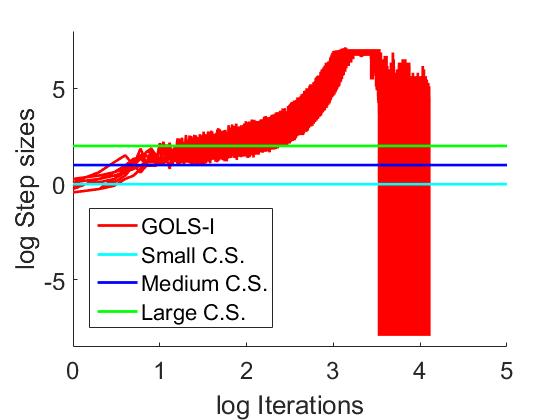}
		\caption{$M = 400$}
	\end{subfigure}%
	
	\caption{Log training error, log training loss, log test error and the log of the step sizes as obtained with various batch sizes for the logistic regression problem for the BCWD dataset.}
	\label{fig_cancer}
\end{figure}

Figure~\ref{fig_cancer} shows log training losses and step sizes for the comparison between GOLS-I and fixed step sizes applied to the logistic regression problem with the BCWD dataset over $100,000$ function evaluations, as first demonstrated by \cite{Mahsereci2017a}. In these plots we show all 10 runs for GOLS-I and each constant step size. In terms of constant step sizes, the small step size exhibits slow convergence, the medium step size performs well, and the large step size often leads to divergence. As the batch size increases, the performance of the large constant step size becomes competitive for isolated instances, as observed for $|\mathcal{B}_{n,i}|=100$ and $|\mathcal{B}_{n,i}|=400$. However, overall the large step size has the largest variance.

The log of the training loss for this problem gives a better perspective of the convergence behaviour of GOLS-I compared to constant step sizes. The training classification errors, as shown in Figure~\ref{fig_cancer_comp}, hides the absolute performance of the training algorithm, due to rounding the network output to obtain the classification. By considering the training loss, we can observe the training performance in more detail, ignoring the effect of rounding. For $|\mathcal{B}_{n,i}|=10$, the variance in the computed gradient between batches is high, hindering the performance of GOLS-I in comparison to the small and medium fixed step sizes. As the batch size increases to $|\mathcal{B}_{n,i}| \geq 50$, the quality of the computed gradient improves sufficiently, such that GOLS-I fits the model to the training data up to numerical accuracy within $100,000$ function evaluations. This point occurs earlier in training, as the mini-batch size is increased.

The step size range of 15 orders of magnitude available to GOLS-I is immediately evident in Figure~\ref{fig_cancer}. At the beginning of training, the step sizes are tightly bound. However, as training progresses, the variance in step sizes increases for $|\mathcal{B}_{n,i}| \in \{10,50,100\}$. This occurs as the decision boundary stabilizes in the classification problem. Samples close to the decision boundary exhibit large gradients and prompt small update steps, others are far from the decision boundary and have small gradients, which prompt GOLS-I to compensate and take large step sizes to find a sign change. Depending on the samples in the mini-batch used to construct the search direction, it is also possible to encounter ascent directions. In such cases, GOLS-I decreases the step size until $\alpha_{min}$ is reached. However, in the smooth and continuous case of $M=400$, the only inaccuracies introduced, are due to GOLS-I's inexact step size resolution, causing the step size rise to be rapid and comparatively consistent between runs. The growing step size behaviour is an adjustment of the line search to the magnitude of the gradient decreasing as the algorithm approaches an optimum. From $10,000$ function evaluations onwards, the variance in step size is due to computational inaccuracy as the gradient approaches numerical 0. At the same point, the loss ceases to decrease during training. This example demonstrates that GOLS-I outperforms constant fixed step sizes and that it generalizes naturally from highly discontinuous, stochastic loss functions to smooth loss functions, with performance increasing as the mini-batch size increases.

\subsection{The MNIST Dataset with NetI and NetII}

\begin{figure}[h]
	\centering
	\begin{subfigure}{0.3\textwidth}
		\centering 
		\includegraphics[width=0.99\linewidth]{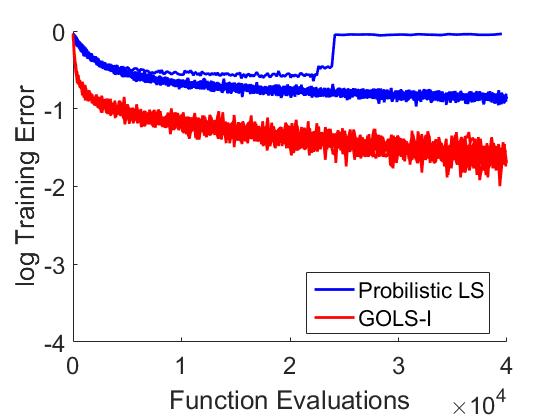}
	\end{subfigure}%
	\begin{subfigure}{0.3\textwidth}
		\centering 
		\includegraphics[width=0.99\linewidth]{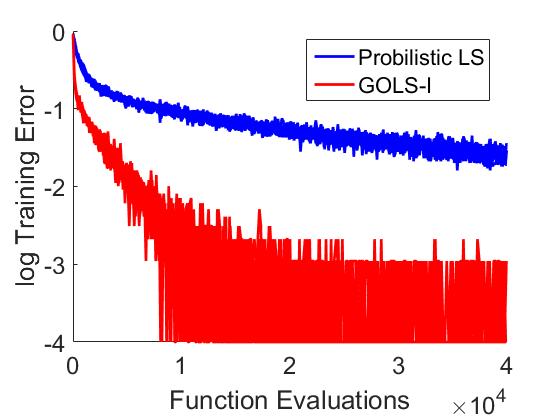}
	\end{subfigure}%
	\begin{subfigure}{0.3\textwidth}
		\centering 
		\includegraphics[width=0.99\linewidth]{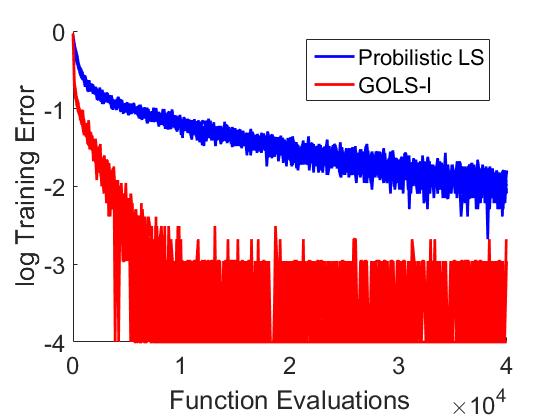}
	\end{subfigure}%
	
	\begin{subfigure}{0.3\textwidth}
		\centering 
		\includegraphics[width=0.99\linewidth]{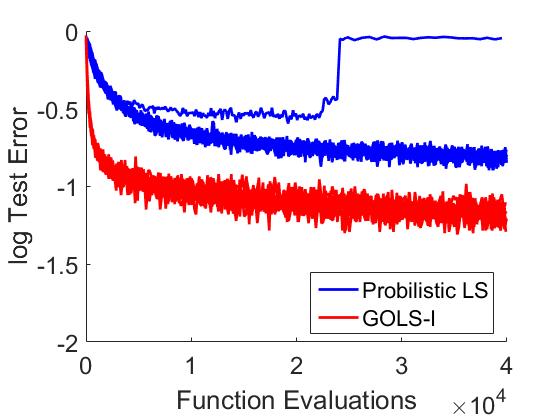}
	\end{subfigure}%
	\begin{subfigure}{0.3\textwidth}
		\centering 
		\includegraphics[width=0.99\linewidth]{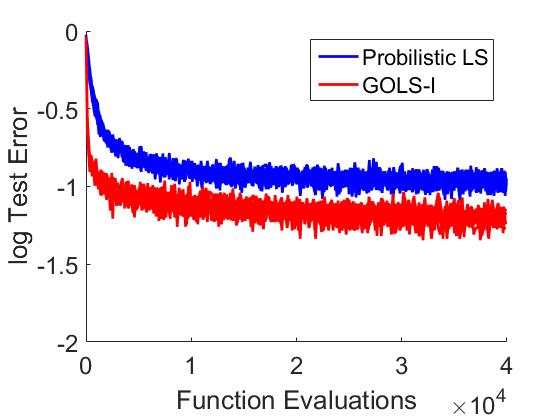}
	\end{subfigure}%
	\begin{subfigure}{0.3\textwidth}
		\centering 
		\includegraphics[width=0.99\linewidth]{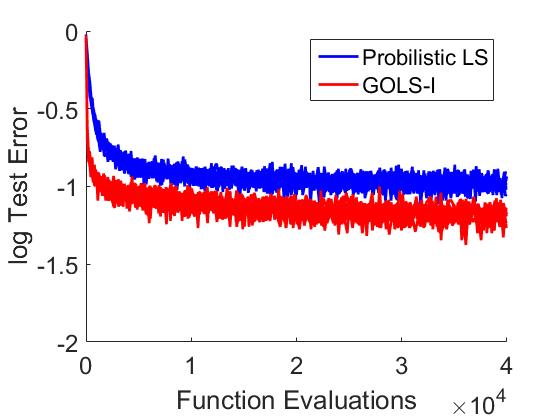}
	\end{subfigure}%
	
	\begin{subfigure}{0.3\textwidth}
		\centering 
		\includegraphics[width=0.99\linewidth]{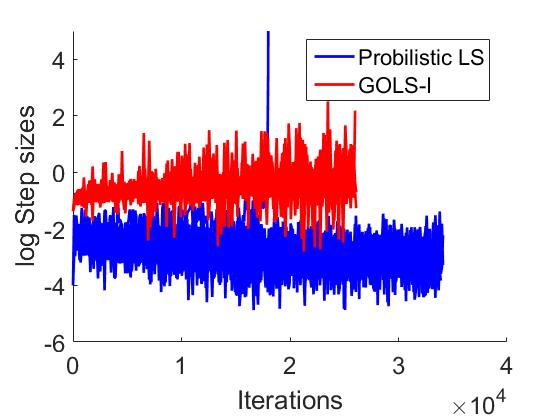}
		\caption{$|\mathcal{B}_{n,i}| = 10$}
	\end{subfigure}%
	\begin{subfigure}{0.3\textwidth}
		\centering 
		\includegraphics[width=0.99\linewidth]{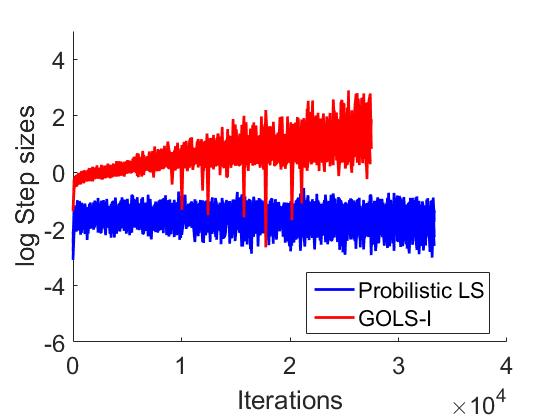}
		\caption{$|\mathcal{B}_{n,i}| = 100$}
	\end{subfigure}%
	\begin{subfigure}{0.3\textwidth}
		\centering 
		\includegraphics[width=0.99\linewidth]{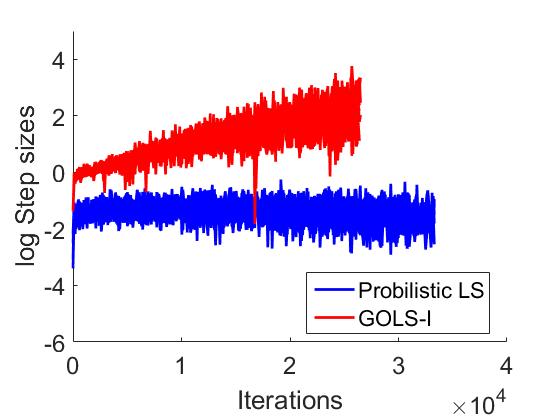}
		\caption{$|\mathcal{B}_{n,i}| = 200$}
	\end{subfigure}%
	
	\caption{log Training error, log training loss, log test error and the log of the step sizes as obtained with various batch sizes for the MNIST dataset with the NetI architecture.}
	\label{fig_MNIST_NI_comp}
\end{figure}

Subsequently, we compare GOLS-I and PrLS as applied to the MNIST dataset with the NetI architecture. The results are given in Figure~\ref{fig_MNIST_NI_comp} for $|\mathcal{B}_{n,i}| \in \{10,100,200\}$. For this example, GOLS-I convincingly outperforms PrLS in both training and test error across all considered mini-batch sizes. PrLS again exhibits divergent behaviour for $|\mathcal{B}_{n,i}| = 10$, while GOLS-I remains stable. There is a significant increase in training performance with GOLS-I, as the mini-batch size increases to $|\mathcal{B}_{n,i}| = 100$. Around 800 function evaluations, some of the first runs obtain zero training classification errors. For $|\mathcal{B}_{n,i}| = 200$, this point is reached around 500 function evaluations. Though training with PrLS also improves as the mini-batch size increases, the average training error reaches a minimum of the order $1 \cdot 10^{-2}$.

As was the case with the BCWD dataset, GOLS-I computes larger step sizes than PrLS while remaining stable during training, which shows that GOLS-I's step sizes are representative of the problem, while PrLS remains conservative. For MNIST on NetI, the number of iterations performed by GOLS-I is visibly less than those of PrLS, indicating that PrLS more readily accepts the initial guess, while not refining the step size to determine a larger step size along the search direction. There is also a difference in step size behaviour between PrLS and GOLS-I: The step sizes determined by PrLS stagnate and even tend to decrease as training progresses, while GOLS-I's step sizes increase during training, which is a function of decreasing gradient magnitudes closer to an optimum. Another noteworthy observation is that both GOLS-I and PrLS automatically determine larger step sizes at the first iteration, as the mini-batch size increases. This is particularly evident between the beginning step sizes determined for $|\mathcal{B}_{n,i}|=10$ and $|\mathcal{B}_{n,i}|=100$ respectively, which is consistent with the constant step size analyses in Figure~\ref{fig_cancer}, where larger step sizes can be more effective with larger mini-batches. The observations of Figure~\ref{fig_MNIST_NI_comp}(last row) indicate that both GOLS-I and PrLS are able to identify this relationship, albeit that PrLS remains more conservative.

\begin{figure}[h]
	\centering
	\begin{subfigure}{0.3\textwidth}
		\centering 
		\includegraphics[width=0.99\linewidth]{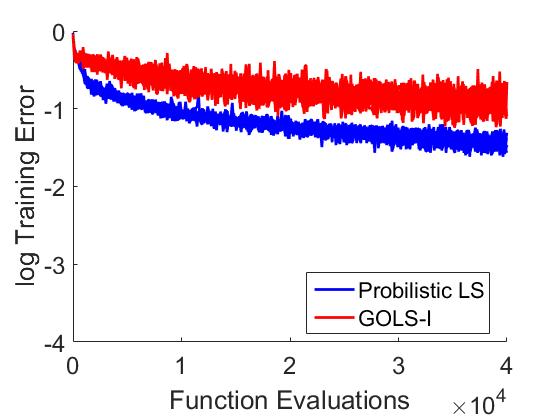}
	\end{subfigure}%
	\begin{subfigure}{0.3\textwidth}
		\centering 
		\includegraphics[width=0.99\linewidth]{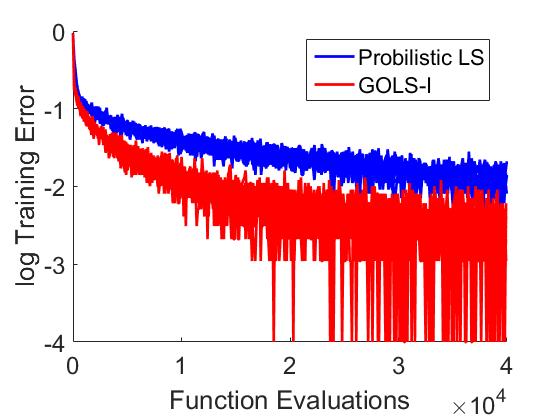}
	\end{subfigure}%
	\begin{subfigure}{0.3\textwidth}
		\centering 
		\includegraphics[width=0.99\linewidth]{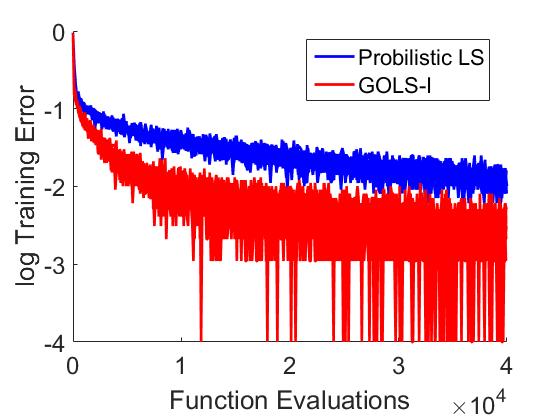}
	\end{subfigure}%
	
	\begin{subfigure}{0.3\textwidth}
		\centering 
		\includegraphics[width=0.99\linewidth]{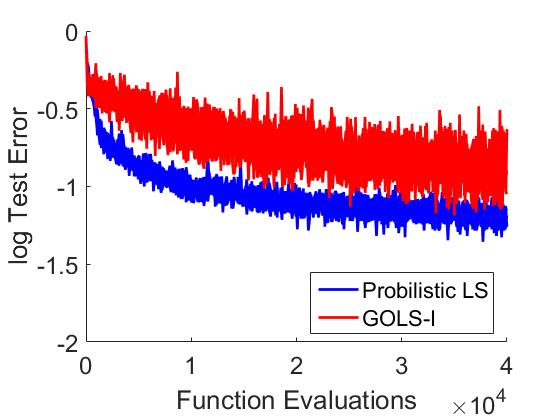}
	\end{subfigure}%
	\begin{subfigure}{0.3\textwidth}
		\centering 
		\includegraphics[width=0.99\linewidth]{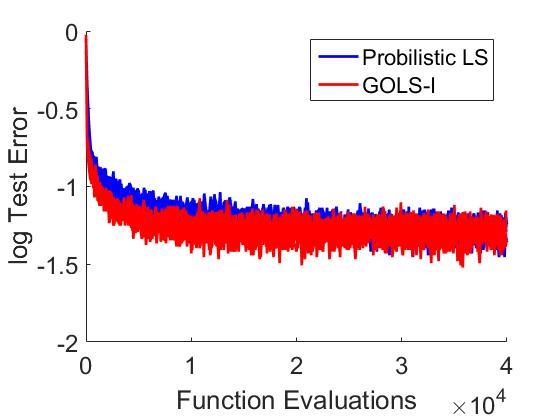}
	\end{subfigure}%
	\begin{subfigure}{0.3\textwidth}
		\centering 
		\includegraphics[width=0.99\linewidth]{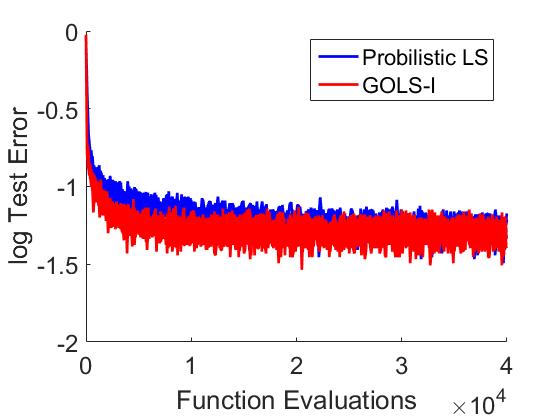}
	\end{subfigure}%
	
	\begin{subfigure}{0.3\textwidth}
		\centering 
		\includegraphics[width=0.99\linewidth]{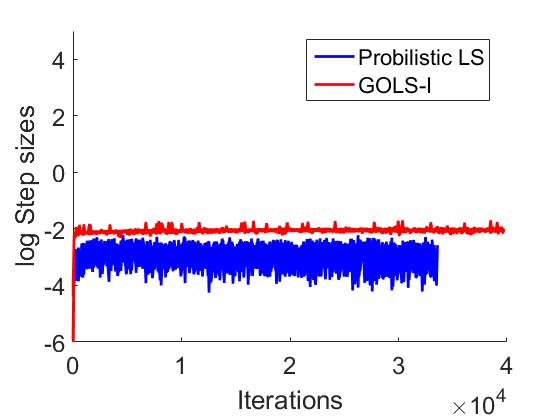}
		\caption{$|\mathcal{B}_{n,i}| = 10$}
	\end{subfigure}%
	\begin{subfigure}{0.3\textwidth}
		\centering 
		\includegraphics[width=0.99\linewidth]{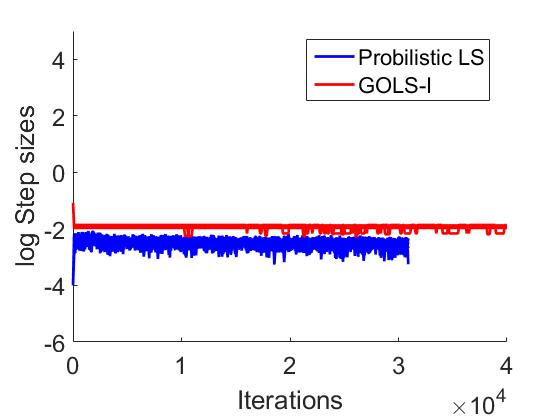}
		\caption{$|\mathcal{B}_{n,i}| = 100$}
	\end{subfigure}%
	\begin{subfigure}{0.3\textwidth}
		\centering 
		\includegraphics[width=0.99\linewidth]{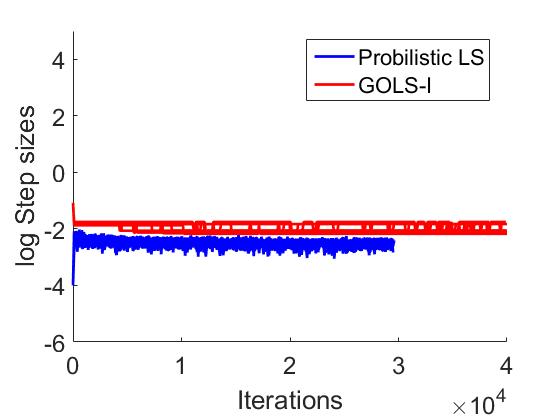}
		\caption{$|\mathcal{B}_{n,i}| = 200$}
	\end{subfigure}%
	
	\caption{Log training error, log training loss, log test error and the log of the step sizes as obtained with various batch sizes for the MNIST dataset with the NetII architecture.}
	\label{fig_MNIST_NII_comp}
\end{figure}

However, this conservatism can have potential benefits, as is demonstrated in Figure~\ref{fig_MNIST_NII_comp} by the training runs conducted on MNIST with NetII for $|\mathcal{B}_{n,i}|=10$. Here PrLS has superior performance to GOLS-I, since GOLS-I often accepts the initial step size, evidenced by the number of iterations performed almost matching the number of function evaluations. Since this trend is the same over the different batch sizes, this indicates that the gradient norms of the problem are reasonably consistent over $|\mathcal{B}_{n,i}|$. In general, GOLS-I again determines larger step sizes than PrLS. However, since the quality of the search direction drops with decreasing $|\mathcal{B}_{n,i}|$, the larger step sizes determined by GOLS-I for $|\mathcal{B}_{n,i}|=10$ lead to noisy, slow training in comparison to PrLS. The smaller step sizes determined by PrLS for $|\mathcal{B}_{n,i}|=10$ results in more stable, consistent training. It is likely, that the additional information used by PrLS works to its advantage over GOLS-I, when the quality of information is poor.

However, as the batch size increases to $|\mathcal{B}_{n,i}|\in \{100,200\}$ the search direction and directional derivative quality improve sufficiently, such that GOLS-I again convincingly outperforms PrLS in training. In turn, the test classification errors are comparable between the two line search methods, indicating that both aid in constructing trained models, albeit that GOLS-I trains the model more efficiently. It is important to note, that the step sizes determined by both algorithms are smaller than those of NetI, and remain constant throughout training for both line searches. This indicates that both line search methods are adapting the estimated step sizes to the information presented by the loss function. The comparative number of iterations performed by PrLS is lower for NetII than for NetI, indicating that PrLS required more function evaluations to estimate the step sizes for each iteration, instead of accepting the initial guess, which was the dominant mode of operation for GOLS-I in this example.

\subsection{The CIFAR10 dataset with NetI and NetII}

\begin{figure}[h!]
	\centering
	\begin{subfigure}{0.3\textwidth}
		\centering 
		\includegraphics[width=0.99\linewidth]{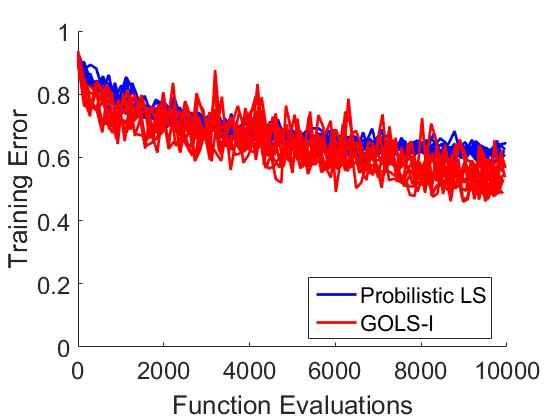}
	\end{subfigure}%
	\begin{subfigure}{0.3\textwidth}
		\centering
		\includegraphics[width=0.99\linewidth]{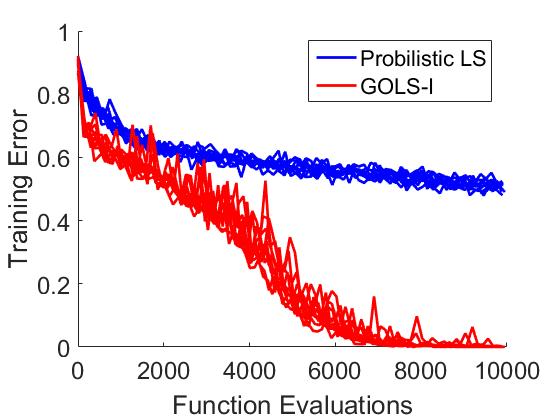}
	\end{subfigure}%
	\begin{subfigure}{0.3\textwidth}
		\centering
		\includegraphics[width=0.99\linewidth]{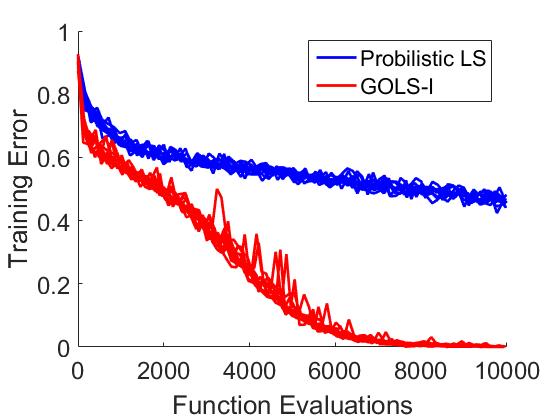}
	\end{subfigure}%
	
	\begin{subfigure}{0.3\textwidth}
		\centering
		\includegraphics[width=0.99\linewidth]{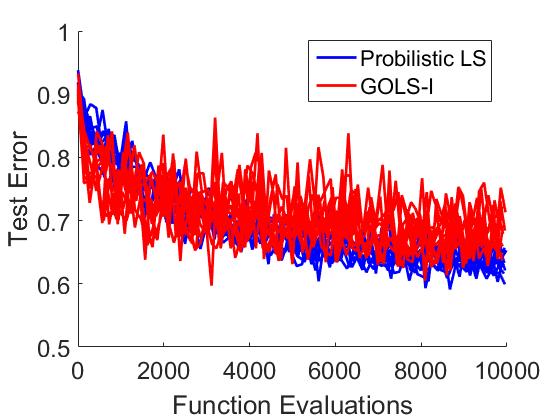}
	\end{subfigure}%
	\begin{subfigure}{0.3\textwidth}
		\centering 
		\includegraphics[width=0.99\linewidth]{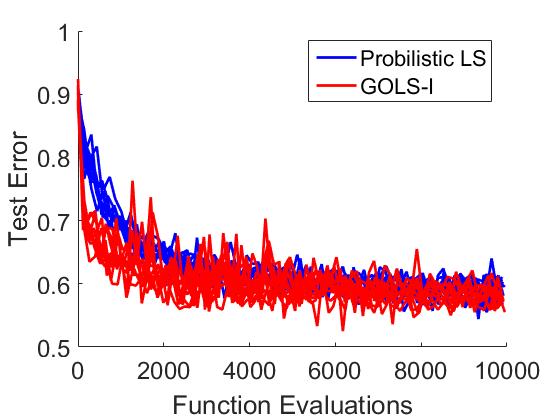}
	\end{subfigure}%
	\begin{subfigure}{0.3\textwidth}
		\centering 
		\includegraphics[width=0.99\linewidth]{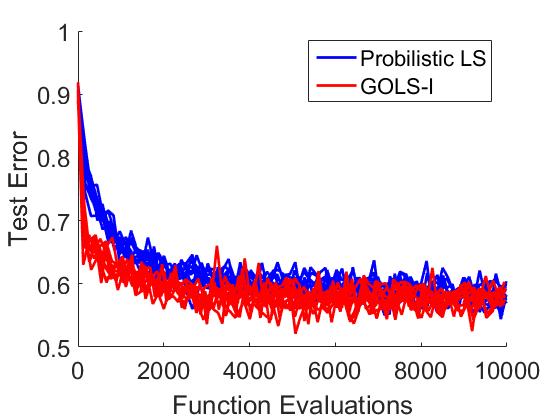}
	\end{subfigure}%
	
	\begin{subfigure}{0.3\textwidth}
		\centering
		\includegraphics[width=0.99\linewidth]{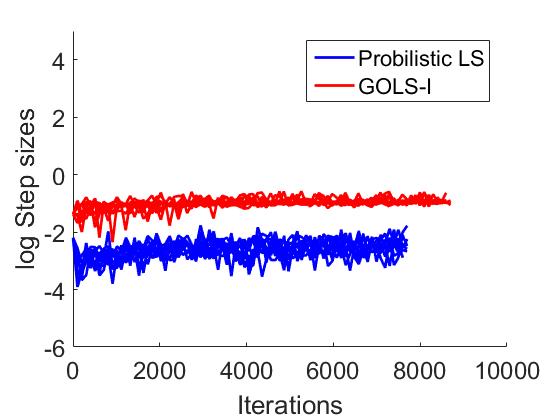}
		\caption{$|\mathcal{B}_{n,i}| = 10$}
	\end{subfigure}%
	\begin{subfigure}{0.3\textwidth}
		\centering 
		\includegraphics[width=0.99\linewidth]{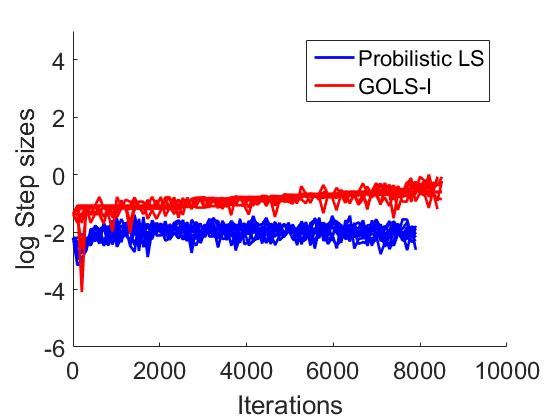}
		\caption{$|\mathcal{B}_{n,i}| = 100$}
	\end{subfigure}%
	\begin{subfigure}{0.3\textwidth}
		\centering 
		\includegraphics[width=0.99\linewidth]{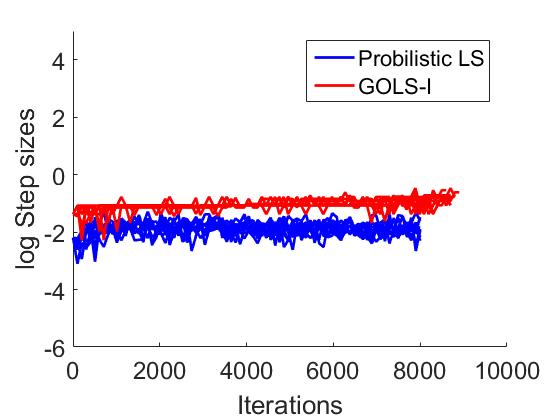}
		\caption{$|\mathcal{B}_{n,i}| = 200$}
	\end{subfigure}%
	
	\caption{Training error, test error and the log of step sizes as obtained with various batch sizes for the CIFAR10 Dataset, as used with the NetI architecture. The training and test errors are shown on a linear scale to allow comparison with results presented by \cite{Mahsereci2017a}.}	
	\label{fig_c_NI}
\end{figure}

The training and test classification errors, with corresponding step sizes for CIFAR10 with NetI are shown in Figure~\ref{fig_c_NI}. To be consistent with results presented in \cite{Mahsereci2017a}, the training and test error plots are plotted on a linear scale for all CIFAR10 analyses. For $|\mathcal{B}_{n,i}|=10$, GOLS-I marginally outperforms PrLS in training. This is again due to GOLS-I determining larger step sizes than PrLS, which also results in noisier training and higher test errors, due to large updates with inconsistent search directions. Again, the conservative step sizes of PrLS aids stability while training with the large discontinuities present for $|\mathcal{B}_{n,i}|=10$. However, as mini-batch size increases to $|\mathcal{B}_{n,i}|\in \{100,200\}$, the performance of GOLS-I again improves significantly, while the improvement for PrLS is marginal. For $|\mathcal{B}_{n,i}| = 100$ GOLS-I trains NetI to within 10\% training classification error within 5,000 function evaluations, while PrLS manages 50\% at best after 10,000 function evaluations. It is to be noted, that the high test classification errors of the CIFAR10 analyses are due to the training dataset containing only Batch1, a $5^{th}$ of the total available training data for this problem. Therefore, the networks used for this dataset overfit readily, and do not generalize well.
In terms of step sizes, PrLS again remains conservative in comparison to GOLS-I. However, the difference between step sizes determined by GOLS-I and PrLS decreases as the mini-batch size increases. Another notable trend, is that the number of iterations performed by PrLS increases as a function of mini-batch size. This indicates, that the initial accept condition of PrLS is more often satisfied, when more information is available. PrLS also exhibits a notable jump of an order of magnitude in step size between $|\mathcal{B}_{n,i}|=10$ to $|\mathcal{B}_{n,i}|=100$. Though it is clear, that PrLS adjusted to the increased quality of information afforded by larger mini-batch sizes, it does not lead to the same increase in performance compared to GOLS-I.

\begin{figure}[h!]
	\centering
	\begin{subfigure}{0.3\textwidth}
		\centering 
		\includegraphics[width=0.99\linewidth]{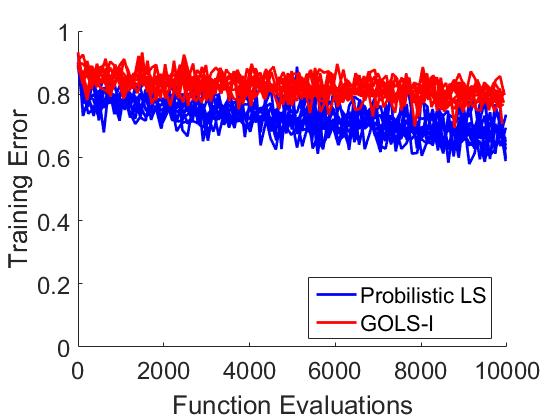}
	\end{subfigure}%
	\begin{subfigure}{0.3\textwidth}
		\centering
		\includegraphics[width=0.99\linewidth]{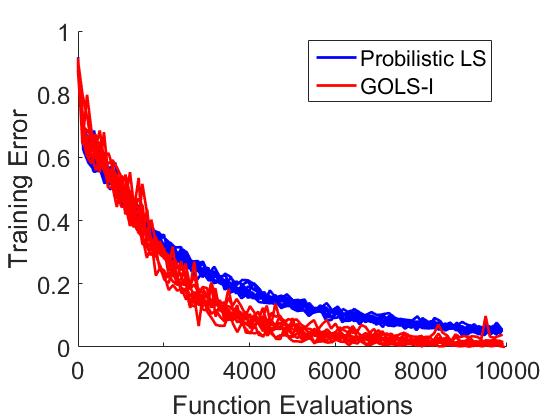}
	\end{subfigure}%
	\begin{subfigure}{0.3\textwidth}
		\centering
		\includegraphics[width=0.99\linewidth]{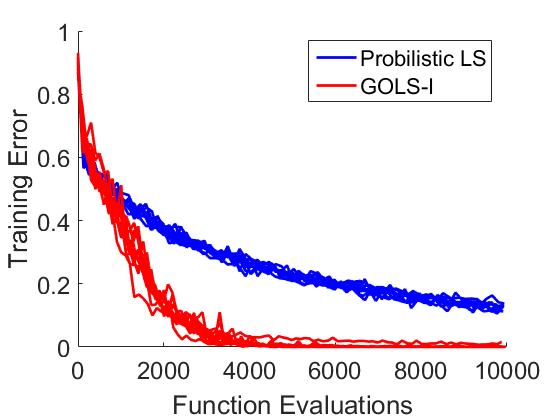}
	\end{subfigure}%
	
	\begin{subfigure}{0.3\textwidth}
		\centering
		\includegraphics[width=0.99\linewidth]{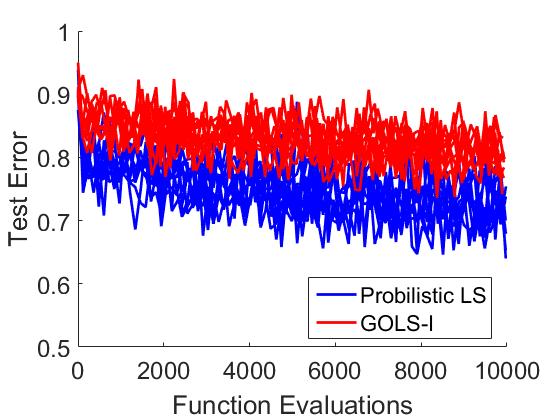}
	\end{subfigure}%
	\begin{subfigure}{0.3\textwidth}
		\centering 
		\includegraphics[width=0.99\linewidth]{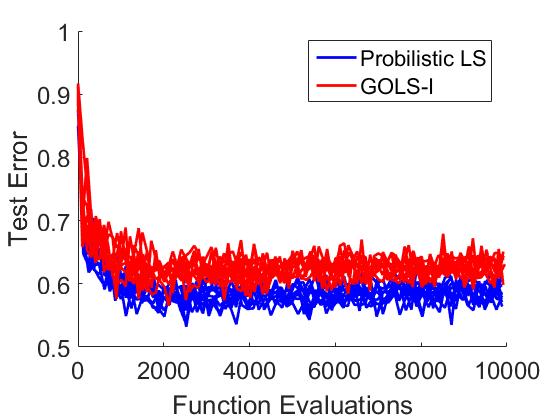}
	\end{subfigure}%
	\begin{subfigure}{0.3\textwidth}
		\centering 
		\includegraphics[width=0.99\linewidth]{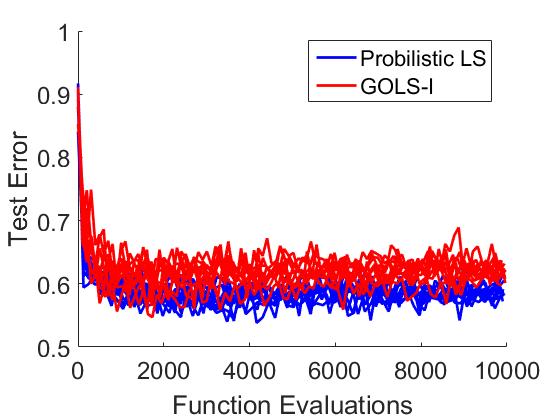}
	\end{subfigure}%
	
	\begin{subfigure}{0.3\textwidth}
		\centering
		\includegraphics[width=0.99\linewidth]{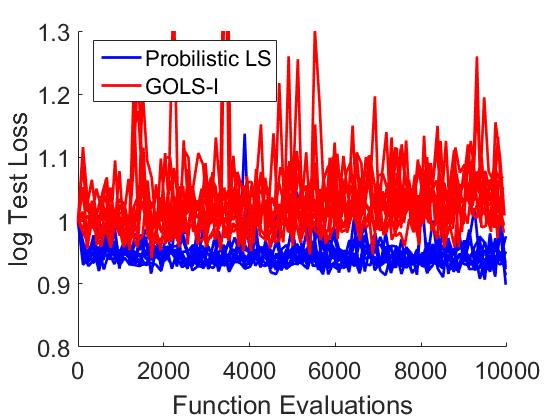}
	\end{subfigure}%
	\begin{subfigure}{0.3\textwidth}
		\centering 
		\includegraphics[width=0.99\linewidth]{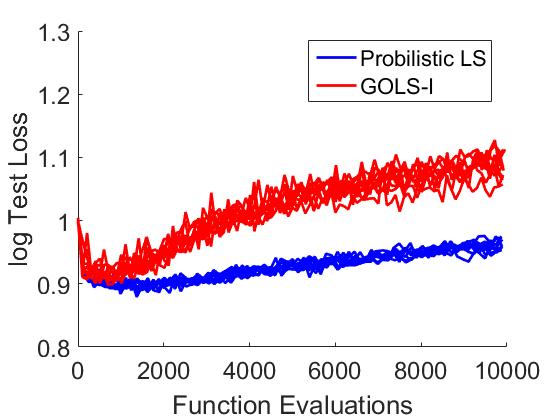}
	\end{subfigure}%
	\begin{subfigure}{0.3\textwidth}
		\centering 
		\includegraphics[width=0.99\linewidth]{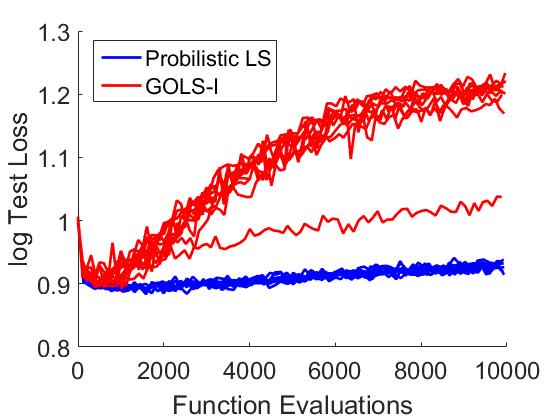}
	\end{subfigure}%
	
	\begin{subfigure}{0.3\textwidth}
		\centering
		\includegraphics[width=0.99\linewidth]{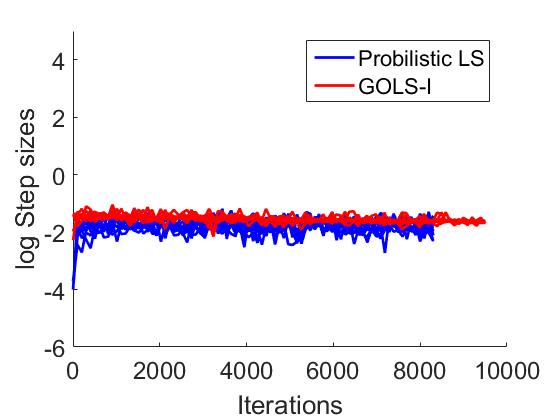}
		\caption{$|\mathcal{B}_{n,i}| = 10$}
	\end{subfigure}%
	\begin{subfigure}{0.3\textwidth}
		\centering 
		\includegraphics[width=0.99\linewidth]{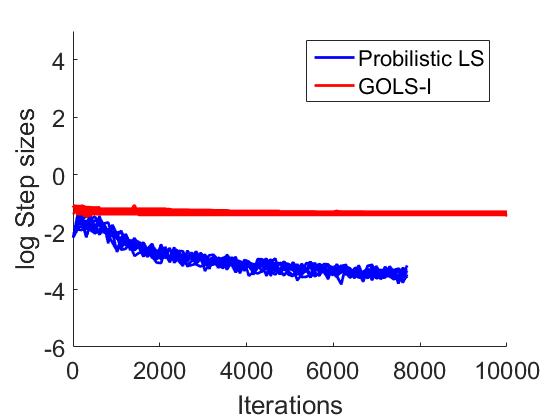}
		\caption{$|\mathcal{B}_{n,i}| = 100$}
	\end{subfigure}%
	\begin{subfigure}{0.3\textwidth}
		\centering 
		\includegraphics[width=0.99\linewidth]{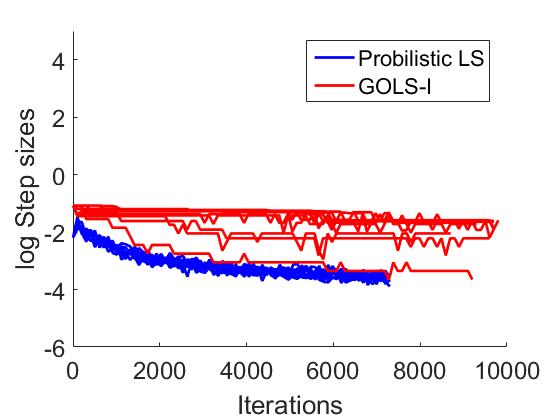}
		\caption{$|\mathcal{B}_{n,i}| = 200$}
	\end{subfigure}%
	
	\caption{Training error, test error, log of test loss and the log of step sizes as obtained with various batch sizes for the CIFAR10 Dataset, as used with the NetII architecture. The training error is shown on a linear scale to allow comparison to results produced by \cite{Mahsereci2017a}.}	
	\label{fig_c_NII}
\end{figure}

Consider the results for CIFAR10 with NetII, shown in Figure~\ref{fig_c_NII}. Here, the conservatism of PrLS again benefits to training with $|\mathcal{B}_{n,i}|=10$. Though the step sizes determined by GOLS-I and PrLS are less than an order of magnitude apart (the closest seen in our investigations), those of PrLS are still on average smaller than those of GOLS-I. This has the implication, that PrLS outperforms GOLS-I in training and test classification errors on NetII with $|\mathcal{B}_{n,i}|=10$. In turn, there is a significant increase in the performance of GOLS-I with $|\mathcal{B}_{n,i}|\in \{100,200\}$, while simultaneously PrLS remains competitive with GOLS-I for $|\mathcal{B}_{n,i}| \geq 100$. The test error of PrLS is lower than that of GOLS-I, which is related to: 1) the training data, 2) the network architecture, and 3) the rate at which training occurs. Apart from training only with Batch1, which allows overfitting to occur readily with NetI, NetII has significantly higher flexibility due to having 3 hidden layers. This further increases the ability of NetII to overfit to the incomplete training data. The rate at which overfitting occurs is directly correlated to training performance. To highlight this aspect, we include the test loss in the third row of Figure~\ref{fig_c_NII}. With $|\mathcal{B}_{n,i}|\in \{100,200\}$, training occurs so rapidly, that the minimum test loss (the point at which overfitting begins) occurs within the first 1,000 function evaluations. As training continues, the training algorithms quickly move past solutions which generalize well, to solutions which minimize the training loss. Consequently, both the test loss and test classification errors continue to increase. Slower training results in the algorithm remaining around solutions which generalize for a larger number of iterations, resulting in lower test loss and test classification errors. This is the case for PrLS for this example, which trains slowly and therefore delays overfitting.

This observation is supported by considering the step sizes shown in the last row of Figure~\ref{fig_c_NII}. For mini-batch sizes $|\mathcal{B}_{n,i}|\in \{100,200\}$, the step sizes determined by GOLS-I remain constant, while those of PrLS decrease three orders of magnitude from $\alpha_{n,I_n} \approx 1\cdot 10^{-1}$ to $\alpha_{n,I_n} \approx 1\cdot 10^{-4}$ during training. This contributes to the slower training behaviour of PrLS, as well as the less aggressive overfit. Unfortunately, this performance can not be attributed to PrLS adapting to the data imbalance of the problem, since the line search is only exposed to the training data, for which it is tasked with minimizing the error. Since PrLS is less successful than GOLS-I in doing so, its lower test error is purely coincidental, given the unique qualities of the problem. 

The results of $|\mathcal{B}_{n,i}|=200$ show an unexpected slowing of training for PrLS compared to $|\mathcal{B}_{n,i}|=100$. This is confirmed by the test losses, where PrLS overfits faster with $|\mathcal{B}_{n,i}|=100$ than with $|\mathcal{B}_{n,i}|=200$. Conversely, the larger mini-batch size once more favours GOLS-I, resulting in efficient training for $|\mathcal{B}_{n,i}|=200$. This behaviour by PrLS is remarkable, since its step size schedule seems unchanged compared to training with $|\mathcal{B}_{n,i}|=100$. Therefore, the only other factor is the quality of the search direction. We postulate, that the variance in search direction allowed PrLS to overcome local optima more effectively with $|\mathcal{B}_{n,i}|=100$, while $|\mathcal{B}_{n,i}|=200$ causes local optima to be more defined. In combination with its conservative step sizes, this slows the training progress of PrLS. The same occurs to one training run with GOLS-I, which determines much smaller step sizes. However, this occurs only after 1000 iterations, at which point the neural network has already overfit, leading to no obvious deficit in training classification.

In all 5 problems considered, the step sizes were adapted dynamically by both GOLS-I and PrLS. Overall, we find PrLS to be more conservative than GOLS-I in terms of step size magnitude. We find PrLS to be unstable at $|\mathcal{B}_{n,i}| << M$ and full-batch analyses with the BCWD dataset, but superior in training for MNIST and CIFAR10 datasets, when $|\mathcal{B}_{n,i}| = 10$. In the latter case, we postulate that this is due to the use of added information in the form of function value and gradient variance in highly discontinuous loss functions. However, the absolute training performance PrLS with $|\mathcal{B}_{n,i}| = 10$ remains underwhelming in comparison to the performance gain offered by GOLS-I with $|\mathcal{B}_{n,i}| \geq 100$. For $|\mathcal{B}_{n,i}| \in \{100,200\}$, the conservatism of PrLS leads to slower training performance, while GOLS-I gains a significant performance advantage from the increased quality of gradient information, which causes SNN-GPPs to be more localized. Additionally, GOLS-I does not require the evaluation of variance estimates, nor the construction of surrogates at every iteration.



\section{Differences to the original PrLS study}

Considerable effort went into following the information given in \cite{Mahsereci2017a} and implementing PrLS as prescribed, especially with regards to constructing the required loss and gradient variance estimates. However, there were some elusive differences between results presented in \cite{Mahsereci2017a} and our investigations, of which we have not yet found the origin.

Firstly, we were unable to match the resolution obtained in log classification error plots, particularly those of the BCWD dataset problems. This is due to the small number of samples in the dataset. Particularly for the test classification errors, we observe plots with discrete accuracy values, due to the binary nature of individual points being either correctly or falsely classified. Though this is less notable in larger datasets, the same characteristics are present.

Secondly, for the most part, we were not able to reproduce the same training behaviour for PrLS as shown in \cite{Mahsereci2017a}. Their work shows training and test errors that drop rapidly, then plateau in the log domain for most of their investigated training problems. We do not recover this behaviour for PrLS nor for GOLS-I. Instead, we observed linear convergence in the log domain for both methods, which is consistent with theoretical convergence estimates for SGD \citep{Dekel2012,Li2014}.

And lastly, though our results for training CIFAR10 with PrLS most closely match those demonstrated in \cite{Mahsereci2017a}, we observe some inconsistencies. For our implementations of NetI, we obtain clearly inferior training performance with PrLS compared to those shown in \cite{Mahsereci2017a}. Conversely, the training of NetII with PrLS is superior with $|\mathcal{B}_{n,i}|=100$ and competitive with $|\mathcal{B}_{n,i}|=200$ to those presented by \cite{Mahsereci2017a}. By experimentation with our numerical parameters, we noted that the given problems are sensitive to the initial guesses of the neural network weights. However, these are not stated in detail in \citep{Mahsereci2017a}. We therefore suspect, that this might be a potential contributor to some of the encountered discrepancies. We highlight these differences in the interest of reproducible science, not distracting from the PrLS method itself, as it clearly serves its intended purpose in determining step sizes. All parameters were kept constant in this study between PrLS and GOLS-I, with the only difference being the actual line search methods used. Although there are differences between our results, and those obtained for PrLS in \cite{Mahsereci2017a}, the comparisons in our investigations hold for the parameters presented in this paper.

\section{PrLS and GOLS-I: Line searches, not update rules}

\begin{figure}[h!]
	\centering
	\begin{subfigure}{.39\textwidth}
		\centering 
		\includegraphics[width=0.99\linewidth]{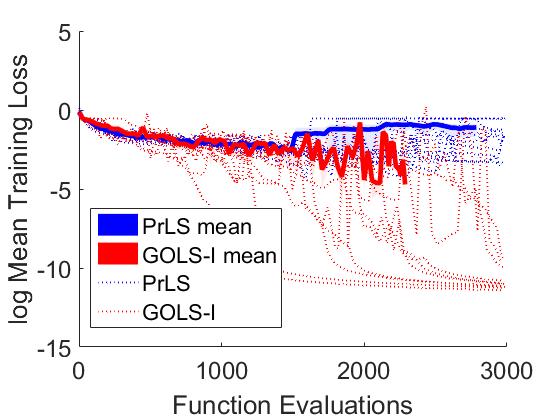}
		\caption{Training loss}
	\end{subfigure}%
	\begin{subfigure}{.39\textwidth}
		\centering
		\includegraphics[width=0.99\linewidth]{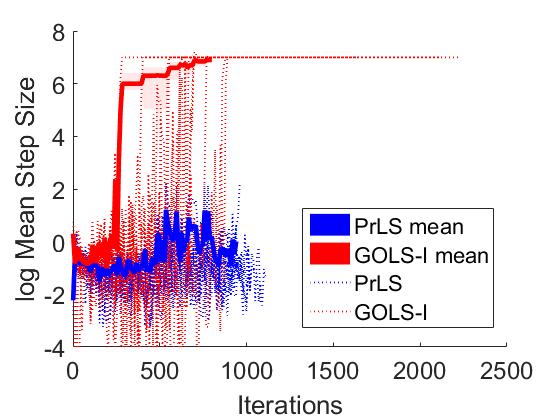}
		\caption{Step size}
	\end{subfigure}%
	
	\begin{subfigure}{.39\textwidth}
		\centering
		\includegraphics[width=0.99\linewidth]{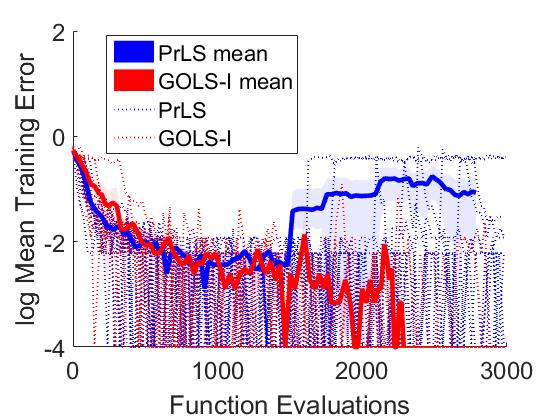}
		\caption{Training classification error}
	\end{subfigure}%
	\begin{subfigure}{.39\textwidth}
		\centering
		\includegraphics[width=0.99\linewidth]{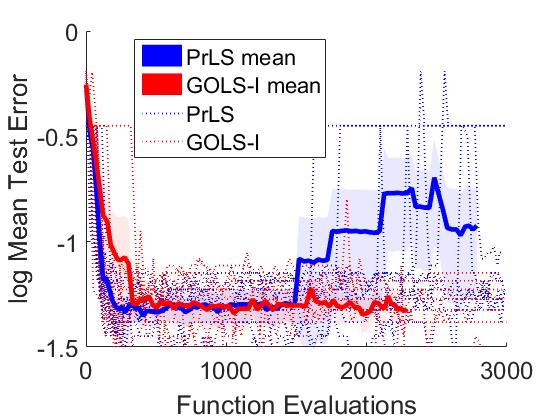}
		\caption{Test classification error}
	\end{subfigure}%
	
	\caption{Training loss, step size, training classification error and test classification error for the Cancer dataset problem with a modified NetPI architecture. We extend NetPI to contain 10 hidden layers, thereby making the training problem more non-linear, which prompts the line search algorithms to perform more function evaluations per iteration.}	
	\label{fig_c_HL10}
\end{figure}

While training MNIST with NetII and CIFAR10 with NetI and NetII, the resulting step sizes for both PrLS and GOLS-I showed that the immediate accept condition was operating more often than the line search method itself. This can be deduced from the number of function evaluations being close to the number of iterations performed. This gives the impression that these line search methods function solely as "update rules", choosing step sizes based on a heuristic, but not actively resolving step sizes. We counter this claim with a demonstration: We select the NetPI architecture with the BCWD dataset and extend the number of hidden layers with 32 nodes from 1 to 10. All other parameters remain the same as those used in Section \ref{sec_BCWD} for NetPI, where training runs are limited to 3,000 function evaluations. We remind the reader, that the first initial guess for each training run is $\alpha_{0,0} = 10^{-4}$ for PrLS and $\alpha_{0,0}=10^{-8}$ for GOLS-I respectively. The large number of hidden layers increases the non-linearity of the model, increasing the complexity of the loss function. To ensure that the loss function is discontinuous, while containing sufficient information, the mini-batch size of $|\mathcal{B}_{n,i}|=100$ is chosen. The resulting training loss, step size, training classification error and test classification error are shown in Figure~\ref{fig_c_HL10}. We show individual results for each of the 10 performed training runs in thin, dotted lines, while highlighting the mean performance over the lowest common number of iterations between runs in a thick, solid line.

The higher non-linearity in the loss function makes optimization within the loss landscape more sensitive to the determined step size. Accuracy of the step size may be more critical in such cases, as the penalty in loss for drastically overshooting an optimum may be more severe. Therefore, the onus is on PrLS and GOLS-I to adjust their respective step sizes to the increased non-linearity in the stochastic loss function. Repeated adjustment of step sizes is indicated by an increased number of function evaluations performed per iteration within the line search. The increased non-linearity decreases the probability of the initial guess from the previous iteration being appropriate for the current iteration, reducing the likelihood of the initial accept conditions being satisfied. 

To shed light on the actions taken by the line searches, we summarize the minimum, maximum and mean number of function evaluations (Fe.) performed per iteration (It.) during training for PrLS and GOLS-I in Table~\ref{tbl_fepit}. The minimum number of Fe./It. is indicative of the immediate accept condition being triggered, while the maximum number of Fe./It. is indicative of the "effort" exerted by the line search to determine the step size for a given iteration. For PrLS, this number is capped to 6(+1) as prescribed by \cite{Mahsereci2017a}. GOLS-I adjusts the step size repeatedly, until either a sign change in directional derivative is found, or one of the min/max step size limits is reached. This allows for a significantly larger number of Fe./It. to be performed at each iteration. Hence, the maximum number of Fe./It. being performed in the first iteration by GOLS-I is 28, as it adjusts the step size from the conservative initial guess of $\alpha_{0,0}=\alpha_{min}=1\cdot 10^{-8}$ to an appropriate magnitude of around $\alpha_{0,I_n}=1\cdot 10^{0}$. In subsequent iterations after initialization, the maximum number of Fe./It. performed by GOLS-I during continued training for this problem was 11. This indicates that both PrLS and GOLS-I were capable of continually adjusting step sizes according to the requirements of the loss landscape. This assertion is also supported by the significant variance in step sizes early on in training. This results in a lower number of iterations performed during the training runs for a maximum 3,000 function evaluations. Generally, the number of Fe./It. drops as training progresses, resulting in lower overall Fe./It. averages. This applies in particular for GOLS-I, where the directional derivatives encountered at the step size upper limit are negative, triggering the initial accept condition.

\begin{table}
	\centering
	\begin{tabular}{|c|c|c|c|}
		\hline 
		 & \textbf{Min Fe./It.} & \textbf{Max Fe./It.} & \textbf{Mean Fe./It.} \\ 
		\hline 
		\textbf{GOLS-I} & 1 & 28 (11) & 1.3 \\ 
		\hline 
		\textbf{PrLS} & 1 & 7 & 2.7 \\ 
		\hline 
	\end{tabular} 
	\caption{Various metrics of function evaluations (Fe.) performed per iteration (It.) during training for PrLS and GOLS-I. Function evaluations performed during an iteration indication of the "effort" exerted by the line search to determine the step size. The maximum number of Fe./It. in PrLS is fixed, which is not the case for GOLS-I. The absolute maximum number of Fe./It. for GOLS-I is 28, which occurs in the first iteration due to a conservative initial guess of $\alpha_{0,0}=1\cdot 10^{-8}$. In subsequent training, the maximum number is 11.}
	\label{tbl_fepit}
	
\end{table}

This investigation demonstrates, that both PrLS and GOLS-I adapt to the loss function characteristics presented to them and can be considered functional line searches in discontinuous loss functions. As is consistent with the previous analyses conducted in this study, GOLS-I shows improved performance over PrLS for the given example. GOLS-I is able to reduce the loss to $1 \cdot 10^{-10}$ for isolated training runs as well as obtain zero training classification errors, which is not the case with PrLS. Instead, PrLS exhibits divergent training behaviour after 1500 function evaluations on average.

Overall, GOLS-I has been shown to be more aggressive than PrLS in terms of the step size magnitudes, which can lead to detrimental performance when small mini-batches are used. However, given the hardware capabilities currently available to machine learning practitioners, it is feasible to implement mini-batch sizes of $|\mathcal{B}_{n,i}|\geq 100$, where GOLS-I has outperformed PrLS for the problem considered.
We also remind the reader, that no variance estimates or surrogates are needed to implement GOLS-I, making its application to existing machine learning technologies less involved and computationally more efficient than PrLS.

\section{Conclusion}

For discontinuous dynamic mini-batch sub-sampled (MBSS) loss functions, we compare the Gradient-Only Line Search that is Inexact (GOLS-I) \citep{Kafka2019jogo}, to the Probabilistic Line Search (PrLS) \citep{Mahsereci2017a} for automatically resolving learning rates. GOLS-I is an intuitive, computationally efficient alternative line search method, which does not require surrogates or function value and gradient estimates, while remaining robust in discontinuous loss functions. Instead of minimizing or finding critical points along descent directions, GOLS-I locates Stochastic Non-Negative Associated Gradient Projection Points (SNN-GPPs). Moving along a 1-D descent direction, SNN-GPPs are identified by sign changes from negative to positive in the directional derivative, thus incorporating second order information representative of a minimum. 

We demonstrate the capabilities of GOLS-I on eight machine learning problems, with five proposed by \cite{Mahsereci2017a}, three adapted from \cite{Prechelt1994}. These include the Breast Cancer Wisconsin Diagnostic (BCWD) dataset with four different architectures, as well as MNIST and CIFAR10 each implemented with a shallow and deep network architecture. We use these problems to demonstrate that step sizes can be efficiently determined for stochastic gradient descent with a line search (LS-SGD) using GOLS-I. GOLS-I adaptively determines step sizes that can vary over 15 orders of magnitude, i.e. from a minimum step size of $\alpha_{min}=10^{-8}$ to a maximum of $\alpha_{max}=10^7$. In our experiments, GOLS-I demonstrated training performance that is competitive to superior to that of a manually tuned constant step size. As training progressed, GOLS-I was able to dynamically re-adjust the step size, depending on the characteristics of the loss function.

We have shown GOLS-I to outperform the probabilistic line search (PrLS) in training when mini-batches are sufficiently large, while the performance of PrLS was complementary to GOLS-I, in that it performed best with small mini-batch sizes. The combination of PrLS being more conservative than GOLS-I in the step sizes, as well as the added information used by PrLS (in the form of loss function and gradient variance estimates) on average makes PrLS more robust than GOLS-I for the MNIST and CIFAR10 problems with mini-batch sizes $|\mathcal{B}_{n,i}|=10$. For the BCWD dataset, PrLS exhibited divergent behaviour with $|\mathcal{B}_{n,i}|=10$ and some instances full-batch training. However, using mini-batch sizes of $|\mathcal{B}_{n,i}|=10$ results in slow training overall. Current computational resources support mini-batch sizes of sufficient size ($|\mathcal{B}_{n,i}|\geq 100$), such that gradient information alone, in the form of the SNN-GPP as used by GOLS-I, is sufficient to enable effective step sizes to be determine in dynamic MBSS loss functions. In such cases, GOLS-I comprehensively outperformed PrLS in this study, which consists largely of problems and architectures proposed by the authors of PrLS. This makes GOLS-I a credible alternative to determining learning rates in dynamic mini-batch sub-sampled loss functions, which leads to curiosity regarding the feasibility of incorporating GOLS-I into other traditional mathematical programming methods and popular neural network training algorithms.

\section*{Acknowledgements}

This work was supported by the Centre for Asset and Integrity Management (C-AIM), Department of Mechanical and Aeronautical Engineering, University of Pretoria, Pretoria, South Africa. We would also like to thank NVIDIA for sponsoring the Titan X Pascal GPU used in this study.

\newpage


\newpage

\bibliography{BibC2}
\bibliographystyle{plainnat}

\end{document}